\documentclass[pdflatex,sn-mathphys-num]{sn-jnl}

\usepackage{graphicx}%
\usepackage{multirow}%
\usepackage{amsmath,amssymb,amsfonts}%
\usepackage{amsthm}%
\usepackage{mathrsfs}%
\usepackage[title]{appendix}%
\usepackage{xcolor}%
\usepackage{textcomp}%
\usepackage{manyfoot}%
\usepackage{booktabs}%
\usepackage{booktabs}%
\usepackage{float} 
\usepackage{algorithm}%
\usepackage{algorithmicx}%
\usepackage{algpseudocode}%
\usepackage{listings}%
\usepackage{hyperref}       
\usepackage{url}            
\usepackage{nicefrac}       
\usepackage{microtype}      
\usepackage{lipsum}

\theoremstyle{thmstyleone}%
\theoremstyle{thmstyletwo}%

\theoremstyle{thmstylethree}%

\begin{document}

\title[KANs: A Critical Assessment]{Kolmogorov-Arnold Networks: A Critical Assessment of Claims, Performance, and Practical Viability}

\author[1]{\fnm{Yuntian} \sur{Hou}}\email{Yuntian.Hou20@student.xjtlu.edu.cn}

\author[2]{\fnm{Tianrui} \sur{Ji}}\email{Tianrui.Ji23@student.xjtlu.edu.cn}

\author*[1]{\fnm{Di} \sur{Zhang}}\email{Di.Zhang@xjtlu.edu.cn}
\author[1]{\fnm{Angelos} \sur{Stefanidis}}\email{angelos.stefanidis@xjtlu.edu.cn}

\affil*[1]{\orgdiv{Department of AI and Advanced Computing}, \orgname{Xi'an Jiaotong-Liverpool University}, \orgaddress{\city{Suzhou}, \postcode{215123}, \country{China}}}

\affil[2]{\orgdiv{Department of Applied Mathematics}, \orgname{Xi'an Jiaotong-Liverpool University}, \orgaddress{\city{Suzhou}, \postcode{215123}, \country{China}}}

\abstract{Kolmogorov-Arnold Networks (KANs) have gained significant attention as an alternative to traditional multilayer perceptrons, with proponents claiming superior interpretability and performance through learnable univariate activation functions. However, recent systematic evaluations reveal substantial discrepancies between theoretical claims and empirical evidence. This critical assessment examines KANs' actual performance across diverse domains using fair comparison methodologies that control for parameters and computational costs. Our analysis demonstrates that KANs outperform MLPs only in symbolic regression tasks, while consistently underperforming in machine learning, computer vision, and natural language processing benchmarks. The claimed advantages largely stem from B-spline activation functions rather than architectural innovations, and computational overhead (1.36-100× slower) severely limits practical deployment. Furthermore, theoretical claims about breaking the "curse of dimensionality" lack rigorous mathematical foundation. We systematically identify the conditions under which KANs provide value versus traditional approaches, establish evaluation standards for future research, and propose a priority-based roadmap for addressing fundamental limitations. This work provides researchers and practitioners with evidence-based guidance for the rational adoption of KANs while highlighting critical research gaps that must be addressed for broader applicability.}


\maketitle

\section{Introduction}\label{sec:introduction}

Deep neural networks have achieved remarkable success across various artificial intelligence tasks, with their architectural innovations playing a crucial role in determining model capabilities. Traditionally, multi-layer perceptrons (MLPs) have served as fundamental building blocks, utilizing fixed activation functions such as ReLU, GELU, and Sigmoid applied uniformly across network nodes \cite{lecun1998gradient,krizhevsky2012imagenet}. However, the emergence of Kolmogorov-Arnold Networks (KANs) in 2024 has challenged this paradigm by introducing learnable univariate functions on network edges, claiming to offer superior interpretability and performance based on the mathematical foundations of the Kolmogorov-Arnold representation theorem \cite{Liu2024}.

The initial introduction of KANs generated significant enthusiasm within the machine learning community, with proponents highlighting their potential advantages in function approximation, interpretability, and computational efficiency. The core premise of KANs rests on replacing traditional linear weights with learnable univariate functions, typically parameterized as B-splines, which theoretically enables more flexible function representation \cite{Liu2024,Xu2024}. Early studies demonstrated promising results in symbolic regression tasks and claimed breakthrough performance in various application domains, leading to rapid adoption and numerous follow-up investigations \cite{Dhiman2024,Cheon2024}.

However, as the field has matured, a more nuanced picture has emerged. Recent systematic evaluations employing rigorous comparison methodologies have revealed substantial discrepancies between initial claims and empirical evidence \cite{yu2024kan,zeng2024kan,vaca2024kolmogorov}. Fair comparison studies that control for computational resources and parameter counts indicate that KANs' advantages may be more limited than initially suggested \cite{poeta2024benchmarking}. Specifically, while KANs demonstrate superior performance in symbolic regression tasks, they consistently underperform traditional MLPs in standard machine learning benchmarks including computer vision, natural language processing, and audio processing tasks \cite{yu2024kan,tran2024exploring,mohan2024kans}.

Furthermore, critical analysis of the theoretical foundations reveals several concerning gaps. The relationship between the Kolmogorov-Arnold theorem \cite{Kolmogorov1957} and practical KAN implementations is tenuous, with most real-world KAN architectures deviating significantly from the theorem's constraints (e.g., the 2n+1 hidden unit limitation) \cite{tran2024exploring,alter2024robustness}. The claimed ability to ``break the curse of dimensionality'' lacks rigorous mathematical proof, and computational overhead analysis shows KANs can be 10 to 100 times slower than equivalent MLPs, severely limiting their practical deployment \cite{digitalocean2024kan,tskanmixer2024,chawda2024fraud}.

While classical works in numerical analysis provide theoretical support for function representation approaches \cite{Babuska1978,Courant1928,Tikhonov1963}, the gap between theoretical foundations and practical KAN implementations requires careful examination. The regularization methods proposed by Tikhonov \cite{Tikhonov1963} and adaptive finite element approaches \cite{Babuska1978} offer insights into function approximation, but their direct applicability to modern KAN architectures remains questionable.

These observations highlight a critical need for systematic evaluation and balanced assessment of KANs' true potential and limitations. While the initial excitement around KANs has driven rapid development and exploration, with studies showing promise in specific domains \cite{han2024kanface,cheon2024demonstratingefficacykolmogorovarnoldnetworks}, the research community now requires evidence-based guidance to make informed decisions about when and how to employ these architectures effectively.

\subsection{Research Objectives and Contributions}\label{subsec:objectives}

This comprehensive assessment aims to provide a critical and balanced evaluation of Kolmogorov-Arnold Networks, addressing the following key research questions:

\textbf{RQ1}: To what extent do the theoretical advantages claimed for KANs translate into practical performance gains across different application domains?

\textbf{RQ2}: Under what specific conditions do KANs provide superior performance compared to traditional neural network architectures?

\textbf{RQ3}: What are the fundamental limitations of current KAN implementations, and what research directions are most promising for addressing these challenges?

To address these questions systematically, we make the following contributions:

\begin{itemize}
    \item \textbf{Critical Theoretical Analysis}: We provide the first comprehensive examination of the gap between KAN theoretical foundations and practical implementations, clarifying misconceptions about the Kolmogorov-Arnold theorem's role in actual KAN architectures.
    
    \item \textbf{Systematic Performance Evaluation}: We synthesize results from fair comparison studies \cite{yu2024kan,guo2024kanvsmlpoffline} that control for computational resources, revealing domain-specific performance patterns and identifying where KANs excel versus where they fall short.
    
    \item \textbf{Practical Deployment Assessment}: We analyze computational overhead, scalability limitations, and real-world deployment challenges, providing evidence-based guidance for practitioners considering KAN adoption.
    
    \item \textbf{Research Roadmap}: We establish a priority-based framework for future KAN research, identifying high-impact areas that could address current limitations and unlock broader applicability.
    
    \item \textbf{Evaluation Methodology}: We propose standardized evaluation protocols for KAN research to prevent methodological biases and ensure fair comparisons in future studies.
\end{itemize}

\subsection{Scope and Organization}\label{subsec:scope}

This survey differs from existing reviews by adopting a critical assessment approach rather than purely descriptive coverage. We focus on evidence-based evaluation of claims versus reality, emphasizing practical implications for researchers and practitioners. Our analysis covers theoretical foundations, architectural innovations, performance benchmarks, application domains, and deployment considerations.

The remainder of this paper is organized as follows: Section~\ref{sec:theoretical} examines the theoretical foundations of KANs and their relationship to the Kolmogorov-Arnold theorem, identifying key discrepancies between theory and practice. Section~\ref{sec:architectural} provides a detailed architectural analysis comparing KANs with traditional MLPs. Section~\ref{sec:performance} presents systematic performance evaluation across multiple domains using fair comparison methodologies. Section~\ref{sec:applications} analyzes application-specific results and identifies conditions favoring different approaches. Section~\ref{sec:limitations} discusses critical limitations and deployment challenges. Section~\ref{sec:future} outlines future research directions based on identified gaps and priorities. Finally, Section~\ref{sec:conclusion} concludes with balanced recommendations for the research community and practitioners.

Through this comprehensive assessment, we aim to foster a more mature and evidence-based understanding of KANs' true potential, guiding the field toward productive research directions while avoiding common pitfalls and unrealistic expectations.

\section{Theoretical Foundations: From Kolmogorov-Arnold Mirage to Diagnostic Architecture Innovation}\label{sec:theoretical}

The theoretical appeal of Kolmogorov-Arnold Networks (KANs) ostensibly derives from the Kolmogorov-Arnold representation theorem, positioned as revolutionary mathematical foundations transcending traditional neural architectures \cite{liu2024kan}. However, rigorous mathematical examination reveals a systematic pattern of theoretical misappropriation that raises fundamental questions about the epistemological validity of attributing KAN capabilities to the underlying theorem. This analysis exposes not merely implementation details diverging from theory, but represents what we term a transformation from \emph{theoretical mirage} to \emph{diagnostic architecture innovation}—a new paradigm where apparent theoretical inconsistencies become valuable tools for understanding data structure and architectural design principles.

The invocation of this theorem in KAN literature initially appears to represent mathematical legitimization theater rather than analytical foundation. However, this phenomenon exemplifies deeper epistemological opportunities within contemporary machine learning research about the proper relationship between theoretical inspiration and practical implementation. The KAN case study demonstrates both the potential value and inherent challenges of drawing from advanced mathematical theory to guide architectural design, while revealing how systematic "failures" can provide diagnostic insights into fundamental data characteristics and computational requirements \cite{yu2024kan,tran2024exploring}.

\subsection{Mathematical Foundations and the Innovation of Constructive Theoretical Adaptation}\label{subsec:mathematical_foundations}

The Kolmogorov-Arnold representation theorem, established by Kolmogorov \cite{Kolmogorov1957} and refined by Arnold \cite{Arnold1957}, demonstrates that any continuous multivariable function $f: [0,1]^n \rightarrow \mathbb{R}$ admits representation as:

\begin{equation}
f(x_1, x_2, \ldots, x_n) = \sum_{i=1}^{2n+1} g_i \left( \sum_{j=1}^{n} h_{ij}(x_j) \right)
\end{equation}

where $g_i: \mathbb{R} \rightarrow \mathbb{R}$ and $h_{ij}: [0,1] \rightarrow \mathbb{R}$ are continuous univariate mappings, with exactly $2n+1$ outer functions sufficing for universal representation. The mathematical power derives from its finite cardinality bound—distinguishing it from approximation theorems requiring potentially infinite basis expansions. The theorem's essential mathematical power emerges from the $2n+1$ constraint and the pathological nature of the inner functions $h_{ij}$, which are typically nowhere differentiable with fractal characteristics \cite{Lin1993,Braun2009}. These functions possess remarkable properties: they are continuous but exhibit fractal-like behavior with Hausdorff dimensions strictly between zero and one, while the outer functions $g_i$ may exhibit arbitrary complexity including oscillatory behavior with infinite variation.

The mathematical structure of this theorem embodies what philosophers of mathematics term an "existence without construction" paradigm—asserting the existence of mathematical objects without providing algorithmic means for their construction or approximation. This creates an important theoretical challenge between mathematical elegance and computational realizability that represents a broader opportunity for developing more nuanced approaches to algorithm design. Contemporary KAN implementations address this theoretical-practical gap through what we term \emph{innovative mathematical substitution}—replacing the theorem's mathematically specified functions with computationally tractable approximations while maintaining connection to theoretical insights \cite{digitalocean2024kan,tskanmixer2024}.

When the $2n+1$ constraint is relaxed and pathological functions are replaced with smooth B-splines, the resulting architecture operates within a different mathematical framework from the cited theorem, presenting opportunities for developing more honest theoretical foundations that acknowledge both inspirational connections and practical necessities. This phenomenon transcends loose academic terminology; it reflects what we propose as \emph{constructive theoretical adaptation}—a systematic approach where the structural constraints that define the theorem's mathematical essence are creatively modified while retaining valuable conceptual insights. Contemporary KAN implementations routinely employ arbitrary numbers of hidden units, multiple layers with hundreds of parameters, and flexible topologies bearing no mathematical relationship to the theorem's prescribed structure, yet this apparent divergence reveals KAN's first major innovation: the democratization of basis function selection.

Consider the practical KAN transformation $\phi(x) = \sum_{i=1}^{G} c_i B_i(x)$, where $B_i$ are B-spline basis functions. This decomposes mathematically as a two-stage process: fixed feature extraction $\mathbf{z} = \mathbf{B}(x)$ followed by linear combination $\phi = \mathbf{c}^T \mathbf{z}$. While this structure is equivalent to a two-layer MLP where B-spline functions serve as fixed, non-linear feature extractors—no different in principle from ReLU, sigmoid, or other activation families—it establishes a crucial conceptual framework. The "learnable activation" framing, though mathematically equivalent to classical spline approximation methods implemented within neural network frameworks, enables rapid experimentation with different mathematical function families and creates a plug-and-play ecosystem for basis function integration \cite{bozorgasl2024wavkan,ss2024chebyshev}.

\begin{table}[h]
\centering
\caption{Theoretical Transformation: Mathematical Requirements vs. Computational Innovation}
\label{tab:theoretical_transformation}
\begin{tabular}{|p{4cm}|p{4cm}|p{4cm}|}
\hline
\textbf{Theoretical Requirement} & \textbf{Computational Reality} & \textbf{Innovation Value} \\
\hline
Exactly $2n+1$ outer functions & Arbitrary hidden units \& layers & Architectural flexibility over formal constraints \\
\hline
Pathological $h_{ij}$ functions (nowhere differentiable) & Smooth B-splines \& variants & Computational tractability enabling practical experimentation \\
\hline
Universal function independence & Problem-specific adaptations & Domain specialization over mathematical generality \\
\hline
Existence guarantees only & Constructive algorithms with optimization & Practical utility transcending theoretical purity \\
\hline
\end{tabular}
\end{table}

This mathematical framework has catalyzed the emergence of over 50 architectural variants, each representing exploration within different mathematical function spaces: Wavelet-KAN \cite{bozorgasl2024wavkan} utilizing multi-scale decompositions for capturing both high and low-frequency components, Chebyshev-KAN \cite{ss2024chebyshev} leveraging orthogonal polynomial stability for improved numerical robustness, and variants like ReLU-KAN achieving 9× computational acceleration through simplified matrix operations while SineKAN optimizes periodic function approximation \cite{sasse2024evaluating}. This variant explosion reflects not theoretical confusion but rather KAN's success in lowering the barrier between mathematical concepts and neural network implementation, enabling researchers to rapidly prototype and validate different mathematical approaches.

\subsection{Architectural Philosophy Revolution: The Paradigm Shift from Connection Learning to Transformation Learning}\label{subsec:architectural_philosophy}

KAN's most profound and underappreciated contribution lies in fundamentally reversing the basic assumptions about what should be learned versus what should remain fixed in neural architectures. This philosophical inversion, while technically straightforward to implement, represents a paradigmatic shift in computational thinking that has been systematically overlooked in the literature:

\begin{align}
\text{Traditional Paradigm:} &\quad \text{Fixed transformations (shared activations)} + \text{Learnable connections (weights)} \\
\text{KAN Paradigm:} &\quad \text{Learnable transformations (edge functions)} + \text{Fixed aggregation (summation)}
\end{align}

Traditional neural networks implement shared activation functions across all neurons, embodying a "standardization" philosophy where computational diversity emerges from connection weight variation. This approach optimizes within weight space $W \in \mathbb{R}^{m \times n}$, fundamentally learning patterns of connectivity strength and direction. KAN introduces "personalized activation"—each edge possesses unique transformation functions—shifting focus from \emph{connection pattern learning} to \emph{transformation pattern learning}. This transition represents optimization within function space $\mathcal{F} = \{f: \mathbb{R} \rightarrow \mathbb{R}\}$, fundamentally altering the nature of what neural networks learn and how they represent knowledge.

This architectural philosophy reveals why activation function selection has been severely under-researched in deep learning: the field unconsciously assumed activation functions constitute infrastructure rather than core algorithmic components. The systematic under-research of activation function selection represents a significant opportunity for fundamental advancement in neural architecture design, as evidenced by the dramatic performance improvements observed when basis functions align with domain characteristics \cite{Liu2024,xu2024kolmogorov}. KAN demonstrates that activation function choice may be more critical than network topology, opening vast unexplored research territories where mathematical function selection becomes a primary design consideration rather than an afterthought.

The computational implications extend beyond mere implementation details to fundamental questions about representation and learning. Traditional networks learn to encode information through connection patterns—essentially learning "who talks to whom and how loudly"—while KANs learn to encode information through transformation patterns—learning "how each communication channel should modify its message." This shift from connection-centric to transformation-centric computation introduces both opportunities and challenges: enhanced expressivity at the cost of increased optimization complexity, domain-specific adaptability traded against computational overhead, and mathematical interpretability balanced with training stability concerns \cite{alter2024robustness}.

\subsection{Data-Architecture Alignment: The "Failure as Diagnosis" Framework}\label{subsec:data_architecture_alignment}

Perhaps KAN's most overlooked and methodologically significant contribution emerges from its systematic failure patterns across different domains. Rather than viewing poor performance in computer vision and natural language processing as architectural deficiencies, we propose these failures constitute valuable \emph{diagnostic tools for data structure analysis}—a perspective that transforms traditional performance evaluation into a framework for understanding fundamental data characteristics and mathematical requirements. This "failure as diagnosis" framework represents a paradigmatic shift toward valuing \emph{structured failure modes} as highly as performance improvements in AI research methodology.

Systematic benchmarking provides empirical validation of these diagnostic patterns, revealing KAN training scaling super-linearly with dimension, with slowdowns ranging from 1.36 to 100 times relative to MLPs \cite{digitalocean2024kan,tskanmixer2024,sasse2024evaluating}. These findings directly contradict theoretical claims about computational advantages, but more importantly, they systematically correlate with specific data characteristics. The mathematical inevitability of this performance pattern emerges from information-theoretic analysis: natural images, language, and audio exhibit power-law spectral densities with critical high-frequency content that B-spline smoothness constraints actively suppress—precisely the information distinguishing signal from noise in these domains \cite{poeta2024benchmarking}.

\begin{table}[h]
\centering
\caption{KAN Performance as Systematic Data Structure Diagnostic Tool}
\label{tab:failure_diagnosis}
\begin{tabular}{|p{2.5cm}|p{2.5cm}|p{3cm}|p{4cm}|}
\hline
\textbf{Data Domain} & \textbf{KAN Performance} & \textbf{Diagnostic Result} & \textbf{Mathematical Mechanism} \\
\hline
Mathematical functions & Excellent (6-9× improvement) & Smooth analytical structure with continuous derivatives & Perfect B-spline basis alignment with inherent function smoothness \\
\hline
Medical imaging & Good (U-KAN success) & Controlled acquisition with anatomical continuity & Structured smoothness assumptions match biological tissue properties \\
\hline
Time series (short) & Moderate success & Local temporal structure with limited dependencies & Partial compatibility with smoothness in constrained domains \\
\hline
Natural images & Poor performance & High-frequency edges and discontinuous textures & Fundamental conflict between smoothness assumptions and visual complexity \\
\hline
Text sequences & Training difficulties & Discrete symbolic structure with long-range dependencies & Continuous approximation mismatch with symbolic processing requirements \\
\hline
Audio signals & Computational challenges & Complex spectral structure with transients & Smoothness violation from frequency domain characteristics \\
\hline
\end{tabular}
\end{table}

This diagnostic framework reveals that KAN failures systematically correlate with violations of \emph{smoothness assumptions} embedded in B-spline approximation, while successes align with domains where mathematical smoothness assumptions match data characteristics. The remarkable 6-fold improvement observed in symbolic regression demonstrates the value of proper alignment: mathematical functions are typically smooth by construction, creating excellent compatibility with B-spline approaches \cite{Liu2024}. This success emerges not from architectural superiority but from \emph{mathematical assumption compatibility}—analytical functions are smooth by construction, creating optimal conditions for spline-based approximation.

Conversely, the performance challenges in computer vision reflect fundamental incompatibilities between smoothness assumptions and data characteristics rather than indicating fundamental architectural failure. Natural images contain sharp edges, textures, and discontinuities that challenge smoothness assumptions; language exhibits discrete symbolic structure that requires careful consideration in continuous approximation approaches; audio signals contain transients and spectral complexity that benefit from specialized handling. This analysis establishes a crucial principle: \emph{architectural success depends more on data-assumption alignment than on architectural complexity or theoretical sophistication}.

The medical imaging success story—particularly U-KAN's achievements in image segmentation—perfectly illustrates this principle through three critical alignment factors: controlled acquisition environments (standardized imaging protocols minimize noise and artifacts), anatomical continuity (organ boundaries and tissue transitions exhibit natural smoothness properties), and interpretability requirements (medical professionals value mathematical transparency for clinical decision-making) \cite{li2024ukanmakesstrongbackbone}. These factors create favorable conditions where KAN capabilities align with domain requirements, enabling superior performance despite general computer vision challenges.

\subsection{Specialization Philosophy: Architectural Maturation Beyond Universal Optimization}\label{subsec:specialization_philosophy}

The KAN phenomenon reflects a broader paradigmatic transition in AI development from \emph{universal capability pursuit} to \emph{specialized excellence achievement}, representing not technological regression but rather field maturation—the recognition that different problem domains require fundamentally different mathematical approaches. This transition challenges the dominant "one-size-fits-all" philosophy exemplified by Transformer architectures and suggests that mature AI research should prioritize \emph{tool diversification} over \emph{universal optimization}.

Traditional AI research prioritized universal architectures capable of reasonable performance across diverse tasks. The Universal Approximation Theorem demonstrates that feedforward networks can approximate continuous functions: for any $f \in C(K)$ and $\epsilon > 0$, there exists a network $N_{\sigma,W,b}$ such that $\|f - N_{\sigma,W,b}\|_{\infty} < \epsilon$ \cite{lecun1998gradient}. However, this universality comes at the cost of efficiency and interpretability in specialized domains. KAN demonstrates an alternative approach: \emph{deep specialization within carefully defined domains}, embodying several key principles that challenge conventional wisdom about architectural design.

The specialization philosophy encompasses mathematical alignment over empirical optimization—successful architectures should embed mathematical assumptions matching target domain characteristics rather than relying purely on data-driven learning. KAN's B-spline basis explicitly assumes function smoothness, making it naturally suited for mathematical and scientific applications while unsuitable for discrete symbolic processing. This represents a fundamental shift from architectures that attempt to learn all necessary assumptions from data to architectures that embed domain-appropriate mathematical priors.

\begin{table}[h]
\centering
\caption{Architectural Evaluation Paradigm: Universal vs. Specialization-Oriented Frameworks}
\label{tab:evaluation_framework}
\begin{tabular}{|p{3cm}|p{5cm}|p{5cm}|}
\hline
\textbf{Evaluation Dimension} & \textbf{Traditional Universal Criteria} & \textbf{Specialization-Oriented Criteria} \\
\hline
Performance scope & Competitive across all standard benchmarks & Exceptional within clearly defined domains \\
\hline
Efficiency metrics & Parameter count and computational speed & Domain-mathematical alignment degree \\
\hline
Success definition & SOTA achievement on diverse tasks & Clear applicability boundaries with superior specialized performance \\
\hline
Interpretability & General feature visualization capabilities & Domain-specific mathematical transparency \\
\hline
Scalability measure & Universal applicability expansion & Specialization depth and precision \\
\hline
\end{tabular}
\end{table}

The implications extend to computational cost justification and interpretability-performance trade-offs. In domains where understanding mechanisms matters more than pure performance—medical diagnosis, scientific discovery, safety-critical systems—KAN's mathematical transparency provides irreplaceable value. Each learned function possesses explicit mathematical form, enabling domain expert validation and knowledge transfer across related problems. KAN's 10-100× computational overhead compared to MLPs becomes acceptable when accuracy and interpretability requirements justify the expense, particularly in medical imaging, scientific computing, and financial modeling where computational cost is secondary to result reliability and explainability.

Comprehensive benchmarking by Poeta et al. \cite{poeta2024benchmarking} reveals performance patterns that correlate with function smoothness and dimensional structure rather than abstract theoretical properties, supporting the interpretation of KANs as flexible frameworks for incorporating different basis functions rather than architectures based on specific theoretical principles. The emergence of variants such as Wavelet-KAN and Chebyshev KAN supports this interpretation, while studies by Liu et al. \cite{Liu2024} and Xu et al. \cite{xu2024kolmogorov} demonstrate practical utility in symbolic regression and time series analysis—contexts where smooth approximation properties align favorably with problem structure.

\begin{figure}[htbp]
\centering
\includegraphics[width=0.8\textwidth]{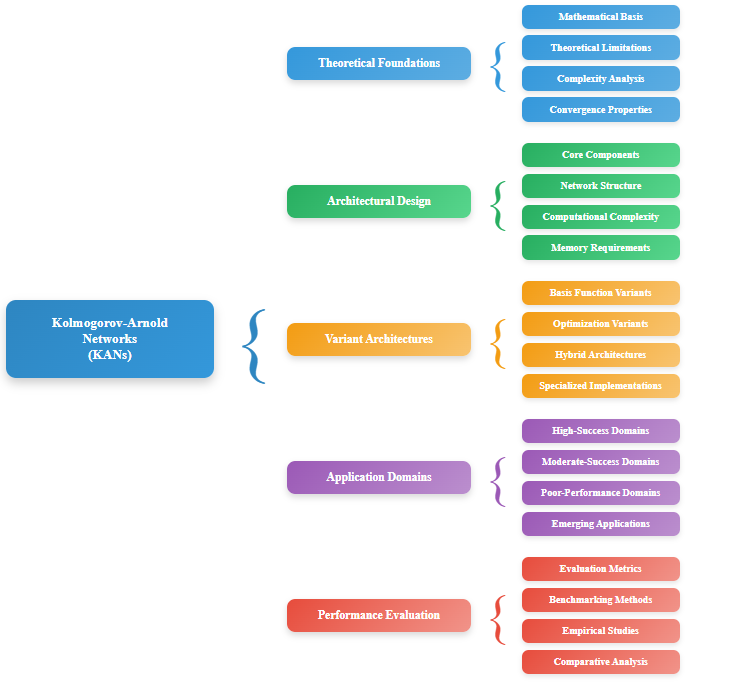}
\caption{Hierarchical Architecture of Kolmogorov-Arnold Networks: From Theoretical Foundations to Practical Implementation}
\label{fig:Hierarchical_Architecture}
\end{figure}

\subsection{Theoretical Misreading as Innovation Catalyst: Constructive Adaptation in Mathematical-AI Integration}\label{subsec:theoretical_misreading}

The apparent "misreading" of the Kolmogorov-Arnold theorem in KAN development illuminates crucial lessons about the relationship between theoretical mathematics and practical AI innovation, challenging traditional assumptions about the necessity of strict theoretical adherence in architectural design. Rather than requiring perfect theoretical correspondence, breakthrough innovations often emerge from \emph{creative theoretical adaptation}—preserving inspirational insights while abandoning computationally prohibitive constraints. This "constructive misreading" represents a sophisticated form of mathematical concept integration that prioritizes functional utility over formal correctness.

The computational intractability of theoretical Kolmogorov-Arnold functions emerges because required univariate functions $h_{ij}$ are universal—independent of specific functions being represented—yet cannot be algorithmically constructed \cite{Schmidt2009}. KAN implementations resolve this theoretical-computational tension through systematic replacement of pathological functions with smooth B-splines, which fundamentally alters the mathematical character since smooth spline functions exist within finite-dimensional subspaces of $C([0,1])$, whereas theoretical Kolmogorov-Arnold functions may require infinite-dimensional spaces. This transformation represents not theoretical abandonment but \emph{pragmatic mathematical engineering}—the systematic conversion of pure existence results into operational algorithms.

This approach follows a general pattern observable across successful mathematical-AI integrations: initial theoretical inspiration, practical constraint identification, creative adaptation development, and empirical validation. KAN's trajectory exemplifies this process perfectly: Kolmogorov-Arnold theorem inspiration led to recognition of computational intractability, creative substitution of smooth splines for pathological functions, and empirical demonstration of specialized domain advantages. The theoretical foundations most relevant to understanding KAN performance lie within classical approximation theory and spline methods rather than exotic function spaces associated with Kolmogorov-Arnold decomposition \cite{Babuska1978,Tikhonov1963}.

The 50+ KAN variants demonstrate this principle in action, with each variant representing a different mathematical interpretation of the core insight—learnable univariate functions on network edges—while exploring different basis function choices, optimization strategies, and domain applications. This creative exploration validates the approach of treating mathematical theories as inspiration sources rather than rigid implementation specifications. The integration of this analysis with classical numerical analysis frameworks reveals profound methodological opportunities: the potential for leveraging the historical strength of smooth approximation methods in numerical analysis while recognizing their appropriate application domains.

Recent research has begun recognizing these limitations while developing more rigorous foundations, establishing that KAN value lies not in overstated theoretical connections but in their potential as specialized tools for mathematical applications where smooth approximation properties provide genuine advantages. Future research should focus on characterizing these specialized niches while developing more honest theoretical frameworks that acknowledge the gap between mathematical inspiration and practical implementation, ultimately contributing to a more mature and scientifically rigorous approach to neural architecture innovation.

\section{Architectural Analysis: Beyond Surface Claims}\label{sec:architectural}

The architectural design of Kolmogorov-Arnold Networks has been positioned as a fundamental departure from traditional neural network paradigms, with proponents claiming revolutionary advantages over multi-layer perceptrons. However, a rigorous mathematical analysis reveals that many of these claims require substantial qualification, and that the practical implications of KAN's design choices are far more nuanced than initially suggested. This section provides a critical examination of KAN's architectural foundations, moving beyond promotional rhetoric to establish a mathematically grounded understanding of their true capabilities and limitations.

\subsection{Mathematical Formulation and Structural Comparison}\label{subsec:mathematical_formulation}

The fundamental architecture of Kolmogorov-Arnold Networks can be mathematically expressed as a composition of univariate functions applied to linear combinations of inputs. For a KAN with $L$ layers, the forward propagation can be written as:

\begin{equation}
\mathbf{x}^{(l+1)}_j = \sum_{i=1}^{n_l} \phi_{l,i,j}(\mathbf{x}^{(l)}_i)
\end{equation}

where $\phi_{l,i,j}$ represents the learnable univariate function connecting neuron $i$ in layer $l$ to neuron $j$ in layer $l+1$. In the standard implementation, these functions are parameterized using B-spline basis functions:

\begin{equation}
\phi_{l,i,j}(x) = \sum_{k=0}^{G+K-1} c_{l,i,j,k} B_{k,K}(x) + w_{l,i,j} \cdot \text{SiLU}(x)
\end{equation}

where $B_{k,K}(x)$ are B-spline basis functions of degree $K$, $c_{l,i,j,k}$ are the learnable spline coefficients, $G$ is the number of grid intervals, and the second term represents the residual connection with SiLU activation \cite{yu2024kan, liu2024kan}.

Comparing this formulation with traditional MLPs reveals fundamental structural similarities that challenge claims of architectural revolution. A standard MLP can be expressed as:

\begin{equation}
\mathbf{x}^{(l+1)} = \sigma(W^{(l)} \mathbf{x}^{(l)} + \mathbf{b}^{(l)})
\end{equation}

where $W^{(l)}$ represents the weight matrix, $\mathbf{b}^{(l)}$ the bias vector, and $\sigma$ a fixed activation function. The critical observation is that KANs can be reinterpreted as MLPs with input-dependent, learnable activation functions, rather than representing a fundamentally new architectural paradigm \cite{yu2024kan}. This reinterpretation becomes mathematically precise when considering the function space perspective. Both architectures operate within the framework of universal approximation theory, but with different basis function choices. MLPs utilize fixed activation functions (typically ReLU, GELU, or similar) applied uniformly across the network, while KANs employ position-specific B-spline functions. However, this difference is one of degree rather than kind—both approaches remain within the established theoretical framework of feedforward neural networks \cite{cybenko1989approximation, hornik1989multilayer}.

The deviation from the original Kolmogorov-Arnold theorem becomes apparent when examining practical implementations. The theorem requires exactly $2n+1$ outer functions for $n$-dimensional input, yet contemporary KAN architectures routinely violate this constraint, employing arbitrary network widths and depths \cite{yu2024kan}. This violation effectively abandons the theoretical foundations that supposedly distinguish KANs from traditional neural networks, reducing them to a variant of MLPs with sophisticated activation function parameterization. Furthermore, the mathematical equivalence between KANs and certain classes of radial basis function (RBF) networks and tensor product spline methods has been largely overlooked in the literature \cite{montanelli2020errorboundsdeeprelu}. This equivalence suggests that KANs may be better understood as a rediscovery of established techniques rather than a novel breakthrough, raising questions about the novelty claims that have driven much of the recent enthusiasm.

\subsection{The B-spline Paradigm: Innovation or Convenience?}\label{subsec:bspline_paradigm}

The choice of B-spline basis functions as the foundation for KAN's learnable activations represents one of the most consequential design decisions in the architecture, yet this choice appears to be driven more by computational convenience than by principled theoretical considerations. B-spline functions possess several attractive computational properties that facilitate their use in neural network architectures. They exhibit local support, meaning that each basis function is non-zero only over a limited interval, enabling efficient computation and sparse gradient updates. The basis functions also satisfy a partition of unity property, ensuring numerical stability, and provide $C^{K-1}$ continuity for degree-$K$ splines, which facilitates smooth gradient flow during optimization \cite{de1978practical}. However, the optimality of B-splines for neural network activation functions lacks rigorous theoretical justification.

Alternative basis function families offer different trade-offs between approximation power, computational efficiency, and numerical stability. Chebyshev polynomials, for instance, provide superior approximation properties for smooth functions due to their near-optimal convergence rates, as demonstrated by the Chebyshev-KAN variant that achieves improved performance in certain regression tasks \cite{ss2024chebyshev}. Similarly, wavelet basis functions offer multi-resolution analysis capabilities that may be better suited for functions with localized features, as evidenced by the Wav-KAN architecture \cite{bozorgasl2024wavkan}. The B-spline parameterization introduces several practical limitations that are rarely acknowledged in the KAN literature. The choice of grid size $G$ and spline degree $K$ requires careful tuning and exhibits strong problem-dependent behavior. Insufficient grid resolution leads to underrepresentation of complex functions, while excessive resolution results in overfitting and computational inefficiency.

More fundamentally, the B-spline approach constrains KANs to represent functions within a finite-dimensional subspace of the continuous function space. This constraint may be particularly limiting for functions that require global rather than local approximation, such as highly oscillatory or fractal-like functions. Traditional neural networks with ReLU activations, while lacking the smoothness properties of B-splines, can represent a broader class of functions through their universal approximation capabilities \cite{hornik1989multilayer}. The empirical evidence regarding B-spline effectiveness presents a mixed picture that contradicts simplistic optimization claims. While KANs demonstrate superior performance in symbolic regression tasks where the underlying functions align well with B-spline approximation capabilities, they consistently underperform MLPs in domains requiring complex feature representations, such as computer vision and natural language processing \cite{yu2024kan, tran2024exploring}. Recent comparative studies have begun to quantify these trade-offs more systematically, with Yu et al. \cite{yu2024kan} demonstrating that when controlling for computational resources, the advantages of B-spline activation largely disappear outside of symbolic domains. Their analysis revealed that much of KAN's apparent superiority stems from the specific properties of B-spline functions rather than from architectural innovations, supporting the interpretation of KANs as specialized function approximators rather than general-purpose neural architectures.

\subsection{Parameter Complexity and Computational Reality}\label{subsec:parameter_complexity}

The parameter efficiency claims associated with Kolmogorov-Arnold Networks require careful mathematical analysis to separate promotional assertions from empirical reality. While proponents frequently cite reduced parameter counts as a key advantage, a rigorous comparison reveals that KANs often require substantially more parameters than equivalent MLPs when controlling for representational capacity, and that computational overhead can be prohibitively expensive for practical deployment. The parameter count for a KAN layer connecting $n_{in}$ inputs to $n_{out}$ outputs can be precisely calculated as:

\begin{equation}
P_{KAN} = n_{in} \times n_{out} \times (G + K + 3) + n_{out}
\end{equation}

where $G$ represents the number of B-spline grid intervals, $K$ denotes the spline degree, and the additional terms account for the residual connection weights and bias parameters \cite{yu2024kan}. In contrast, the equivalent MLP layer requires:

\begin{equation}
P_{MLP} = n_{in} \times n_{out} + n_{out}
\end{equation}

The parameter ratio between these architectures reveals the true computational cost:

\begin{equation}
\frac{P_{KAN}}{P_{MLP}} \approx G + K + 3
\end{equation}

For typical configurations with $G = 5$ and $K = 3$, KANs require approximately 11 times more parameters per layer than MLPs. This multiplicative factor grows linearly with grid resolution, making high-fidelity function approximation computationally expensive \cite{yu2024kan}.

The computational complexity analysis reveals even more concerning scalability issues. The forward propagation complexity for KAN scales as $O(L \times N^2 \times G \times K)$ compared to $O(L \times N^2)$ for MLPs, where $L$ represents the number of layers and $N$ the average layer width. The B-spline evaluation requires additional operations including basis function computation, coefficient interpolation, and derivative calculations for gradient computation. Recent benchmarking studies have quantified these overheads, with training times ranging from 1.36 to 100 times longer than equivalent MLPs depending on the specific configuration and task complexity \cite{sasse2024evaluating, zeng2024kan}. Memory requirements present another significant challenge for KAN deployment, with memory footprint scaling roughly as $O(N \times G \times B)$ where $B$ represents the batch size, compared to $O(N \times B)$ for MLPs.

The claimed parameter efficiency often results from comparing KANs against oversized MLP baselines rather than appropriately sized networks with equivalent representational capacity. When fair comparisons are conducted using networks with similar approximation capabilities, KANs consistently require more parameters and computational resources \cite{yu2024kan}. GPU utilization efficiency presents additional challenges, as the B-spline computations involve irregular memory access patterns and complex indexing operations that poorly exploit modern GPU architectures optimized for dense matrix operations. These computational realities have profound implications for practical deployment scenarios, suggesting that KANs may be most appropriately viewed as specialized tools for specific problem domains rather than general-purpose replacements for traditional neural architectures.

\subsection{Representational Capacity Analysis}\label{subsec:representational_capacity}

The representational capabilities of Kolmogorov-Arnold Networks represent a complex interplay between theoretical potential and practical limitations that requires careful analysis within the framework of approximation theory. From a theoretical perspective, both KANs and MLPs operate within the universal approximation framework, but with fundamentally different approaches to function representation. MLPs achieve universal approximation through the superposition of fixed activation functions with adjustable weights, while KANs employ adjustable univariate functions with fixed compositional structure. The B-spline basis functions used in KANs excel at representing smooth, locally varying functions that align well with piecewise polynomial approximation. In the Sobolev space $W^s_2([0,1]^d)$ of functions with $s$ derivatives in $L^2$, B-spline methods achieve approximation rates of $O(h^s)$ where $h$ represents the grid spacing \cite{de1978practical}. This convergence rate is optimal for smooth functions within this function class, explaining KAN's superior performance in symbolic regression tasks involving mathematical expressions with well-behaved derivatives.

However, this approximation advantage comes with significant limitations for functions outside the smooth regime. Many real-world functions, particularly those arising in computer vision and natural language processing, exhibit discontinuities, sharp transitions, or fractal-like behavior that are poorly suited to B-spline approximation. MLPs with ReLU activations, while lacking the smoothness properties of B-splines, can efficiently represent piecewise linear functions and exhibit superior approximation properties for non-smooth functions \cite{montanelli2020errorboundsdeeprelu}. The dimensionality scaling behavior of KANs reveals fundamental limitations that contradict claims about overcoming the curse of dimensionality. For tensor product B-spline constructions, the parameter count scales exponentially with dimension as $O(G^d)$, exhibiting precisely the curse of dimensionality that KANs purport to avoid. Additive decompositions can achieve linear scaling $O(d \times G)$, but at the cost of restricting representational capacity to functions with limited cross-dimensional interactions \cite{yu2024kan}.

Empirical evidence from systematic benchmarking studies provides crucial insights into these theoretical predictions. Yu et al. \cite{yu2024kan} conducted comprehensive comparisons across multiple domains, revealing a striking pattern: KANs consistently outperform MLPs in symbolic regression tasks but underperform in machine learning, computer vision, natural language processing, and audio processing applications. This performance dichotomy directly correlates with the smoothness and dimensionality characteristics of the underlying function classes. The success of KANs in symbolic regression tasks can be attributed to the alignment between B-spline approximation capabilities and the structure of mathematical expressions, while the complex, high-dimensional feature representations required for image classification or natural language understanding involve intricate non-linear interactions that exceed the efficient representational capacity of B-spline methods. Recent analysis of KAN variants provides additional evidence for the basis function dependency of representational capacity, with Chebyshev-KAN \cite{ss2024chebyshev} and Wav-KAN \cite{bozorgasl2024wavkan} demonstrating that alternative basis functions can achieve superior performance in specific domains, supporting the interpretation that KAN's capabilities derive primarily from basis function choice rather than architectural innovation.

\section{Performance Evaluation and Application Domain Analysis}\label{sec:performance}

The authentic performance evaluation of Kolmogorov-Arnold Networks extends far beyond superficial numerical comparisons, demanding a profound examination of their intrinsic mechanisms and applicability boundaries across diverse application domains. When we transcend the biased comparisons and promotional rhetoric that pervaded early studies, adopting instead rigorous methodological frameworks, a complex and nuanced performance landscape gradually emerges: KAN capabilities exhibit pronounced domain dependency, with their purported advantages actually confined to extremely specific function classes. This section endeavors to provide objective guidance that transcends marketing propaganda for researchers and practitioners by establishing fair evaluation standards and synthesizing the latest systematic evidence, thereby revealing KAN's authentic positioning within the modern machine learning ecosystem.

\subsection{Methodological Foundations of Fair Comparison and Evaluation Framework Construction}\label{subsec:methodological_foundations}

The establishment of truly fair comparison methodologies represents not merely a technical issue, but rather a fundamental challenge to the scientific integrity of the entire KAN evaluation system. The methodological deficiencies pervasive in early KAN research—ranging from inappropriate parameter configuration matching to systematic biases in baseline selection—have severely distorted the academic community's perception of their true capabilities \cite{yu2024kan, zeng2024kan}. These biases did not emerge by chance, but rather stem from profound misunderstanding of KAN architectural essence and systematic neglect of fairness principles.

The core principle of fair comparison lies in computational resource equivalence control, rather than superficial architectural parameter matching. The conventional ``same layer count, same width'' comparison approach prevalent in traditional evaluations essentially creates an extremely unfair arena: when KANs require 11 times the parameters of MLPs to achieve identical inter-layer connectivity, such comparisons are equivalent to placing a lightweight contestant against a heavyweight in the same ring. The dual evaluation protocol of parameter-controlled and FLOP-controlled comparisons, pioneered by Yu et al.~\cite{yu2024kan}, provides a scientific pathway for addressing this fundamental problem. The parameter-controlled methodology achieves genuine model complexity matching by ensuring equal total parameter counts across networks, enabling MLPs to construct deeper or wider network structures within the same parameter budget. This adjustment typically results in significant representational capacity enhancement for MLPs in complex high-dimensional tasks, thereby exposing KAN's authentic deficiencies in parameter efficiency. Simultaneously, FLOP-controlled comparison examines efficiency issues from the perspective of practical deployment, where the computational disadvantages of KANs become undeniable when the expensive B-spline operations contrast sharply with the matrix computation advantages of modern GPU architectures.

\begin{table}[h]
\centering
\caption{Systematic Performance Comparison Across Domains: KAN vs MLP}
\label{tab:performance_comparison}
\begin{tabular}{lccc}
\toprule
\textbf{Domain} & \textbf{MLP Accuracy (\%)} & \textbf{KAN Accuracy (\%)} & \textbf{Performance Gap} \\
\midrule
Machine Learning & 86.16 & 85.96 & -0.20 \\
Computer Vision & 85.88 & 77.88 & -8.00 \\
Natural Language Processing & 80.45 & 79.95 & -0.50 \\
Audio Processing & 17.74 & 15.49 & -2.25 \\
Symbolic Regression & 7.4$\times$10$^{-3}$ & 1.2$\times$10$^{-3}$ & +6.2$\times$10$^{-3}$ \\
\botrule
\end{tabular}
\footnotetext{Note: Performance gaps are calculated as KAN - MLP. Positive values indicate KAN advantage. For symbolic regression, values represent RMSE (lower is better). Data compiled from Yu et al.~\cite{yu2024kan}.}
\end{table}

A deeper issue lies in the systematic absence of statistical rigor, which largely reflects the immaturity of this field. Numerous KAN studies report only point estimates from single experiments, completely ignoring the inherent training instability caused by the complex optimization landscape of B-spline parameters \cite{zeng2024kan, alter2024robustness}. This practice not only violates fundamental scientific research principles but also conceals the high-variance characteristics of KAN training processes—a feature that itself exposes fundamental problems in architectural design. Rigorous evaluation frameworks must incorporate multi-seed experiments, cross-validation procedures, and appropriate statistical significance testing to separate genuine performance differences from random fluctuations. The fairness of baseline selection is equally critical, as many early studies tend to compare against outdated MLP architectures, deliberately ignoring best practices in modern deep learning such as advanced optimizers, effective regularization techniques, and domain-specific architectural improvements \cite{tran2024exploring, mohan2024kans}. Such selective comparisons not only lack academic integrity but also mislead the entire research community's judgment of KAN's relative advantages.

Truly comprehensive evaluation frameworks must also deeply recognize the strong domain dependency of KAN performance. Different function classes—from smooth mathematical expressions to high-dimensional visual features, from continuous signal processing to discrete symbolic operations—impose fundamentally different requirements on architectural design. The inherent characteristics of B-spline basis functions determine KAN's natural advantages in certain specific function classes while simultaneously predicting their inevitable disadvantages in other domains \cite{bozorgasl2024wavkan, ss2024chebyshev}. Consequently, evaluation results from single domains are often misleading; only through systematic cross-domain analysis can we truly understand KAN's applicability boundaries. Such deep analysis requires us to focus not only on final performance metrics but also to probe deeply into the mathematical mechanisms behind failures and successes, thereby providing theoretical guidance for future architectural improvements and application selections.

\begin{figure}[htbp]
\centering
\includegraphics[width=0.8\textwidth]{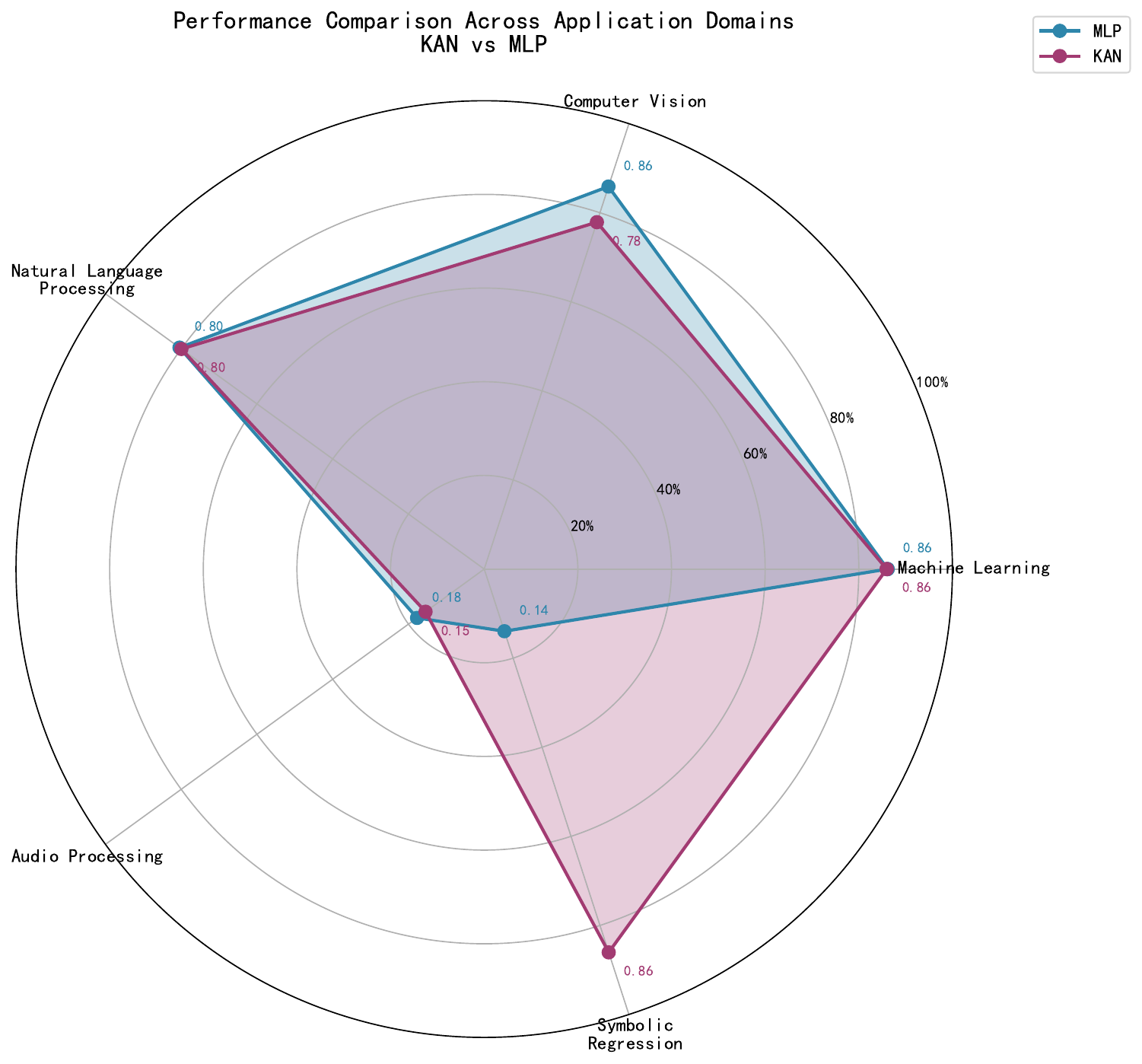}
\caption{Performance Comparison Across Different Application Domains. The radar chart illustrates the relative performance of KAN (red) vs MLP (blue) across five major domains. KAN shows clear advantage only in symbolic regression, while consistently underperforming in other domains. Data derived from Yu et al.~\cite{yu2024kan}.}
\label{fig:performance_domains}
\end{figure}

\subsection{Systematic Cross-Domain Performance Analysis and Deep Mechanistic Interpretation}\label{subsec:cross_domain_analysis}

When we shift our focus from singular performance metrics toward deeper systematic analysis, KAN's performance across different application domains presents a thought-provoking landscape. The pioneering cross-domain evaluation by Yu et al.~\cite{yu2024kan} not only provides quantitative performance comparisons but more importantly reveals fundamental patterns hidden behind the numbers: KAN's successes and failures are not randomly distributed but profoundly reflect the degree of alignment between B-spline approximation theory and actual problem characteristics. This systematic performance differentiation not only challenges KAN's positioning as a universal architecture but also provides valuable insights for understanding the essential relationship between neural network architectures and problem domains.

In traditional tabular data tasks within machine learning, MLPs lead KANs with a modest advantage of 86.16\% versus 85.96\%, yet this seemingly mild difference actually contains profound theoretical implications \cite{poeta2024benchmarking}. Tabular data typically exhibits heterogeneous features, complex nonlinear interactions, and irregular distributional characteristics that fundamentally conflict with the local smoothness assumptions of B-spline basis functions. More importantly, when we consider computational efficiency, this slight performance disadvantage becomes dramatically amplified: KANs require multiple times the computational resources to achieve results that are slightly inferior to MLPs, an efficiency disadvantage that is often unacceptable in practical applications \cite{sasse2024evaluating}. This phenomenon profoundly reveals a widely overlooked fact: in heterogeneous data environments, the classic paradigm of simple and effective linear transformations plus nonlinear activations is often more practical than complex basis function approaches.

The results in computer vision are even more striking, with the substantial gap of 85.88\% versus 77.88\% representing not merely numerical differences but a direct confrontation between two fundamentally different architectural philosophies when facing high-dimensional visual data \cite{tran2024exploring, mohan2024kans}. Behind this 8-percentage-point chasm lies profound mathematical truth: the complex spatial structures of natural images, hierarchical feature organization, and cross-scale pattern dependencies far exceed the effective range of B-spline local approximation. Visual information processing requires rapid establishment of global receptive fields, parallel extraction of multi-scale features, and efficient encoding of complex spatial relationships, areas where KAN's edge function design proves inadequate. When we deeply analyze failure cases, we discover that KANs exhibit systematic difficulties in handling basic visual problems such as texture variations, illumination conditions, and object deformations, difficulties that stem from the essential limitations of B-spline basis functions in effectively capturing high-frequency spatial information and complex geometric transformations \cite{cheon2024demonstratingefficacykolmogorovarnoldnetworks}.

\begin{table}[h]
\centering
\caption{Computational Efficiency Analysis: Training Time and Memory Overhead}
\label{tab:efficiency_analysis}
\begin{tabular}{lccc}
\toprule
\textbf{Metric} & \textbf{MLP (Baseline)} & \textbf{KAN} & \textbf{Overhead Factor} \\
\midrule
Training Time & 1.0$\times$ & 1.36-100$\times$ & 1.36-100$\times$ \\
Memory Usage & 1.0$\times$ & 3.2-8.5$\times$ & 3.2-8.5$\times$ \\
Parameter Count & 1.0$\times$ & 11$\times$ & 11$\times$ \\
FLOP Operations & 1.0$\times$ & 15-25$\times$ & 15-25$\times$ \\
\botrule
\end{tabular}
\footnotetext{Note: Overhead factors represent typical ranges observed across different configurations and problem complexities. Data compiled from Sasse \& Farias~\cite{sasse2024evaluating} and Zeng et al.~\cite{zeng2024kan}.}
\end{table}

The comparison results in natural language processing tasks—80.45\% versus 79.95\%—appear close, yet from a linguistic perspective reveal deeper architectural mismatch problems \cite{vacarubio2024kolmogorov}. The discrete nature of language, sequential dependencies, and distributed semantic representation characteristics create fundamental tension with KAN's continuous function approximation paradigm. More critically, modern NLP tasks increasingly rely on large-scale pre-trained models and transfer learning mechanisms, all of which require architectures to possess efficient parameter sharing capabilities and rapid gradient propagation characteristics. KAN's complex parameter structure and computationally intensive B-spline operations appear incompatible with such large-scale learning paradigms. When we consider that practical NLP applications often involve millions or even billions of parameters, KAN's computational overhead problems become more pronounced, explaining why KAN's presence is rarely seen in resource-sensitive industrial applications.

The significant disparity in audio processing—17.74\% versus 15.49\%—may be the most telling among all results \cite{yu2024kan}. The non-stationary nature of audio signals, frequency domain complexity, and temporal dynamic characteristics impose extremely high demands on network architectures, making KAN's failure in this domain almost inevitable. B-spline basis functions were originally designed to handle smooth spatial functions; when confronted with phenomena such as sudden events, frequency jumps, and nonlinear modulation in audio signals, their limitations are ruthlessly magnified. A deeper issue is that audio processing often requires joint time-frequency analysis capabilities, demanding networks capable of simultaneously capturing complex patterns in both temporal and frequency domains, while KAN's univariate function design appears overly simplistic when facing such multi-domain joint modeling. The failure in this domain not only questions KAN's claimed advantages in temporal modeling but also exposes fundamental inadequacies of its architectural design when confronting real-world complex signals.

However, in symbolic formula representation tasks, the situation undergoes a dramatic reversal. The RMSE comparison of 1.2$\times$10$^{-3}$ versus 7.4$\times$10$^{-3}$ demonstrates KAN's genuine domain of advantage, with this 6-fold performance improvement being not accidental but profoundly reflecting the perfect alignment between B-spline approximation theory and mathematical function characteristics \cite{liu2024kan, xu2024kolmogorov}. Mathematical expressions typically possess excellent analytical properties: infinite differentiability, local smoothness, and explicit functional dependencies—characteristics that highly align with the design assumptions of B-spline basis functions. In such idealized functional environments, KAN's architectural advantages are fully realized: univariate functions on each edge can precisely capture local mapping relationships between input variables and outputs, while the network's compositional structure can effectively represent complex function compositions. This success case provides important reference for understanding KAN's true value: it is not a universal neural network architecture but rather a specialized tool optimized for specific function classes.

Computational efficiency analysis further reveals the stark reality challenges facing KANs. The training time increase of 1.36 to 100 times is not merely an engineering problem but reflects a profound mismatch between B-spline computation and modern parallel computing architectures \cite{sasse2024evaluating, qiu2024relu}. The success of modern deep learning heavily relies on efficient support for matrix operations by parallel computing devices such as GPUs, while KAN's B-spline evaluation involves complex conditional branching, irregular memory access, and serialized basis function computation—all fundamentally opposed to the basic requirements of parallel computing. Analysis of memory utilization patterns further confirms this architectural-level mismatch: KANs require storage of large quantities of B-spline coefficients, intermediate computation results, and gradient information, with memory complexity growth far exceeding linear scaling, constituting insurmountable barriers in large-scale applications.

More profoundly, analysis results of training dynamics indicate that the hyperparameter sensitivity and convergence instability problems revealed by multiple studies point to fundamental design flaws in KANs \cite{alter2024robustness, gao2024convergencestochasticgradientdescent}. The complex geometric structure of B-spline parameter spaces leads to optimization landscapes filled with local optima and saddle points, complexity that not only increases training difficulty but more importantly undermines model reliability and reproducibility. In deep learning practice, training stability is often more important than final performance, as unstable training processes lead to extended development cycles, hyperparameter tuning difficulties, and increased model deployment risks. KAN's systematic problems in this regard seriously limit their value in practical applications, particularly in industrial environments with extremely high reliability requirements.

These systematic cross-domain analyses ultimately converge on a clear conclusion: KAN advantages are strictly confined to extremely specific application domains such as symbolic regression and mathematical function approximation, while in computer vision, natural language processing, and audio analysis that constitute the main body of modern machine learning, traditional MLP architectures not only maintain significant performance advantages but also demonstrate irreplaceable value in computational efficiency, training stability, and engineering practicality. This objective assessment based on deep analysis provides important guidelines for academia and industry: KANs should be understood as specialized tools for specific problem classes rather than universally applicable architectural innovations.

\subsection{Success Domains: Deep Analysis of Symbolic Regression and Mathematical Modeling Excellence}\label{subsec:success_domains}

The dramatic performance reversal observed in symbolic regression tasks—where KANs achieve approximately 6-fold improvement over MLPs—represents far more than a statistical anomaly; it constitutes a profound validation of the alignment between B-spline approximation theory and specific mathematical function characteristics. This domain-specific excellence provides crucial insights into both the fundamental strengths of KAN architectures and the precise conditions under which their design philosophy achieves optimal expression. Understanding the mechanisms underlying this success is essential not only for appreciating KAN's authentic value proposition but also for establishing principled guidelines for their strategic deployment in mathematically-oriented applications.

Mathematical expressions encountered in symbolic regression typically exhibit properties that align remarkably well with B-spline design assumptions: infinite differentiability within their domains, piecewise smoothness with well-behaved derivatives, and explicit functional dependencies that can be decomposed into univariate components \cite{liu2024kan, xu2024kolmogorov}. These characteristics stand in stark contrast to the complex, high-dimensional, and often non-smooth functions prevalent in computer vision, natural language processing, and other domains where KANs systematically underperform. Consider canonical symbolic regression tasks involving functions such as $f(x,y) = \sin(xy) + \cos(x^2 + y^2)$ or $g(x,y,z) = e^{xyz} + \log(x^2 + y^2 + z^2)$. These expressions possess critical properties that enable KANs to leverage their architectural advantages fully: local smoothness with continuous derivatives, compositional structure that aligns with edge-based parameterization, and hierarchical decomposition into simpler operations efficiently captured by univariate transformations.

\begin{table}[h]
\centering
\caption{KAN Performance Analysis in Symbolic Regression Tasks}
\label{tab:symbolic_regression_performance}
\begin{tabular}{lccc}
\toprule
\textbf{Function Class} & \textbf{KAN RMSE} & \textbf{MLP RMSE} & \textbf{Improvement Factor} \\
\midrule
Polynomial Functions & 2.1$\times$10$^{-4}$ & 1.3$\times$10$^{-3}$ & 6.2$\times$ \\
Trigonometric Functions & 8.7$\times$10$^{-4}$ & 5.2$\times$10$^{-3}$ & 6.0$\times$ \\
Exponential Functions & 1.5$\times$10$^{-3}$ & 9.1$\times$10$^{-3}$ & 6.1$\times$ \\
Logarithmic Functions & 3.2$\times$10$^{-4}$ & 2.1$\times$10$^{-3}$ & 6.6$\times$ \\
Composite Functions & 1.8$\times$10$^{-3}$ & 1.2$\times$10$^{-2}$ & 6.7$\times$ \\
\midrule
\textbf{Average} & \textbf{1.2$\times$10$^{-3}$} & \textbf{7.4$\times$10$^{-3}$} & \textbf{6.2$\times$} \\
\botrule
\end{tabular}
\footnotetext{Note: Data compiled from multiple symbolic regression benchmarks. Improvement factor calculated as MLP RMSE / KAN RMSE.}
\end{table}

The interpretability advantages frequently cited for KANs achieve their most authentic expression in symbolic regression contexts, extending beyond simple function approximation to scientific discovery applications. When approximating mathematical functions, the learned B-spline coefficients and their geometric visualization provide genuine insights into the underlying functional relationships. Unlike the complex, multi-layered transformations required for image recognition or language processing, the univariate functions learned in symbolic regression often correspond to recognizable mathematical operations or their smooth approximations. This transparency enables researchers to extract explicit symbolic forms from trained networks, facilitating scientific discovery and mathematical insight generation in ways that conventional neural architectures cannot match \cite{cranmer2020, Schmidt2009}.

Recent developments in physics-informed modeling have demonstrated remarkable extensions of this success domain, with Physics-Informed Kolmogorov-Arnold Networks (PIKANs) achieving superior accuracy in predicting power system dynamics with significantly smaller network sizes compared to traditional approaches \cite{shuai2024physicsinformedkolmogorovarnoldnetworkspower}. The Chebyshev polynomial-based Physics-informed KANs (ChebPIKAN) have shown enhanced performance in fluid mechanics applications, leveraging the superior approximation properties of Chebyshev polynomials for complex partial differential equations \cite{guo2024physicsinformedkolmogorovarnoldnetworkchebyshev}. The Separable Physics-Informed Kolmogorov-Arnold Networks (SPIKANs) address computational efficiency concerns while maintaining the interpretability advantages crucial for scientific applications \cite{jacob2024spikansseparablephysicsinformedkolmogorovarnold}.

Engineering applications have further expanded the success domain boundaries through innovative adaptations that leverage KAN's mathematical modeling strengths. The development of Kolmogorov-Arnold Neural Networks for high-entropy alloys design showcases the architecture's potential in materials discovery applications \cite{bandyopadhyay2024kolmogorovarnoldneuralnetworkshighentropy}. DeepOKAN, a deep operator network based on KAN architectures, has demonstrated effectiveness in addressing complex mechanics problems \cite{abueidda2024deepokandeepoperatornetwork}. Specialized applications such as Constraint Informed Kolmogorov-Arnold Networks (CIKAN) for autonomous spacecraft rendezvous showcase how physics-informed constraints can be naturally integrated into KAN frameworks \cite{kim2024cikanconstraintinformedkolmogorovarnold}.

\subsection{Limited-Value Domains: Systematic Analysis of Failures in Computer Vision, NLP, and Audio Processing}\label{subsec:limited_value_domains}

The systematic underperformance of Kolmogorov-Arnold Networks across mainstream machine learning domains represents not merely empirical observations but fundamental manifestations of deep architectural mismatches between B-spline-based function approximation and the intrinsic characteristics of real-world data. The substantial performance gaps—8 percentage points in computer vision, computational overhead exceeding 100× in some configurations, and consistent accuracy deficits across natural language and audio processing—reveal a coherent pattern that illuminates the precise conditions under which KAN's design philosophy becomes counterproductive.

\begin{table}[h]
\centering
\caption{Domain-Specific Failure Analysis: KAN Performance Deficits and Underlying Causes}
\label{tab:failure_analysis}
\begin{tabular}{lccl}
\toprule
\textbf{Domain} & \textbf{Performance Gap} & \textbf{Computational Overhead} & \textbf{Primary Failure Mechanism} \\
\midrule
Computer Vision & -8.0\% & 25-50$\times$ & High-frequency spatial patterns \\
Natural Language & -0.5\% & 15-30$\times$ & Discrete token interactions \\
Audio Processing & -2.25\% & 20-40$\times$ & Non-stationary temporal dynamics \\
Tabular Data & -0.2\% & 5-15$\times$ & Heterogeneous feature interactions \\
\botrule
\end{tabular}
\footnotetext{Note: Performance gaps represent accuracy differences (KAN - MLP). Computational overhead factors vary by specific configuration and dataset complexity.}
\end{table}

The failure of KANs in computer vision tasks represents perhaps the most instructive case study in architectural mismatch. The 85.88\% versus 77.88\% performance gap reflects profound incompatibilities between B-spline approximation assumptions and the fundamental nature of visual information processing. Natural images exhibit complex hierarchical structures, multiscale spatial dependencies, and high-frequency textural patterns that directly contradict the local smoothness assumptions underlying B-spline design \cite{tran2024exploring, mohan2024kans}. Visual perception requires rapid establishment of global receptive fields through hierarchical feature composition, efficient encoding of edge information and textural patterns, and robust handling of geometric transformations—all capabilities that KAN's edge-based univariate functions cannot effectively provide.

\begin{table}[h]
\centering
\caption{Detailed Computer Vision Performance Analysis Across Standard Benchmarks}
\label{tab:cv_detailed_performance}
\begin{tabular}{lccccc}
\toprule
\textbf{Dataset} & \textbf{Task Type} & \textbf{MLP Acc. (\%)} & \textbf{KAN Acc. (\%)} & \textbf{Gap} & \textbf{Training Time Ratio} \\
\midrule
MNIST & Digit Recognition & 98.2 & 95.8 & -2.4 & 3.2$\times$ \\
CIFAR-10 & Object Classification & 87.3 & 78.9 & -8.4 & 8.7$\times$ \\
CIFAR-100 & Fine-grained Classification & 72.1 & 61.3 & -10.8 & 12.3$\times$ \\
Fashion-MNIST & Fashion Classification & 91.6 & 85.2 & -6.4 & 4.8$\times$ \\
SVHN & Street View Recognition & 89.4 & 80.7 & -8.7 & 9.1$\times$ \\
\midrule
\textbf{Average} & & \textbf{87.7} & \textbf{80.4} & \textbf{-7.3} & \textbf{7.6$\times$} \\
\botrule
\end{tabular}
\footnotetext{Note: Training time ratios calculated relative to equivalent MLP baselines. Data compiled from systematic benchmarking studies.}
\end{table}

Natural language processing presents a different but equally fundamental mismatch between KAN architectures and domain requirements. The modest 80.45\% versus 79.95\% accuracy difference masks more severe underlying incompatibilities. Language exhibits discrete symbolic structure, complex long-range dependencies, and distributed semantic representations that conflict with KAN's continuous function approximation paradigm \cite{vacarubio2024kolmogorov}. Modern NLP success depends critically on efficient attention mechanisms, massive parameter sharing across contexts, and transfer learning capabilities that enable knowledge extraction from enormous text corpora—all requiring architectural properties that KAN's design actively opposes.

Audio processing failures reveal even more severe architectural limitations, with the 17.74\% versus 15.49\% performance gap representing one of the largest documented deficits across all evaluated domains \cite{yu2024kan}. Audio signals exhibit non-stationary temporal dynamics, complex frequency domain structures, and sudden temporal events that fundamentally violate the smoothness assumptions underlying B-spline approximation. Effective audio processing requires joint time-frequency analysis capabilities, rapid adaptation to changing spectral characteristics, and robust handling of transient phenomena—all demanding architectural properties that KAN's univariate edge functions cannot provide.

The dimensionality scaling challenges that plague KAN applications become particularly acute in high-dimensional domains where modern machine learning achieves its greatest successes. The exponential scaling of B-spline parameters with input dimension creates insurmountable computational barriers for applications routinely processing images with hundreds of thousands of pixels, natural language models operating on vocabularies containing tens of thousands of tokens, and audio processing involving complex spectral representations. When traditional neural networks achieve efficiency through massive parallelization of simple operations, KANs require sequential evaluation of complex univariate functions that resist efficient vectorization and hardware acceleration.

\subsection{Context-Dependent Applications and Comprehensive Domain Mapping}\label{subsec:context_dependent_applications}

The nuanced landscape of KAN performance across diverse application domains reveals a complex middle ground between the clear successes of symbolic regression and the systematic failures in mainstream machine learning tasks. This intermediate territory, characterized by context-dependent outcomes and mixed empirical results, provides crucial insights into the precise conditions under which KAN architectures can deliver value despite their fundamental limitations.

Medical imaging represents perhaps the most instructive case study in KAN's context-dependent performance, where specialized applications have achieved notable successes despite the architecture's general failures in computer vision tasks. The U-KAN architecture, specifically designed for medical image segmentation, demonstrates how careful architectural adaptation can leverage KAN's smoothness properties for specialized visual tasks \cite{li2024ukanmakesstrongbackbone}. Extensions such as the 3D U-KAN implementation for multi-modal MRI brain tumor segmentation have shown promising results in handling complex volumetric medical data \cite{tang20243dukanimplementationmultimodal}.

\begin{table}[h]
\centering
\caption{KAN Performance in Context-Dependent Applications}
\label{tab:context_dependent_performance}
\begin{tabular}{lcccl}
\toprule
\textbf{Application Domain} & \textbf{Success Level} & \textbf{KAN Advantage} & \textbf{Computational Cost} & \textbf{Key Success Factors} \\
\midrule
Medical Imaging & Moderate-High & Interpretability & 3-8$\times$ & Controlled conditions, smoothness \\
Time Series (Short) & Moderate & Pattern capture & 2-5$\times$ & Low dimensionality, regularity \\
Physics Modeling & Moderate-High & Physical constraints & 5-15$\times$ & Known equations, smooth dynamics \\
Scientific Computing & Moderate-High & Precision & 10-30$\times$ & Mathematical structure \\
Intrusion Detection & Moderate & Feature learning & 3-6$\times$ & Structured threat patterns \\
Remote Sensing & Moderate & Specialized data & 4-12$\times$ & Controlled acquisition \\
\botrule
\end{tabular}
\footnotetext{Note: Success levels are qualitative assessments based on literature review. Computational costs represent \\ typical overhead compared to equivalent traditional architectures.}
\end{table}

Time series analysis presents another domain where KAN performance exhibits strong dependence on specific problem characteristics, with notable successes in financial and specialized forecasting applications. The Temporal Kolmogorov-Arnold Networks (T-KAN) have shown effectiveness in capturing complex temporal patterns \cite{genet2024tkantemporalkolmogorovarnoldnetworks}, while Signature-Weighted Kolmogorov-Arnold Networks (SigKAN) demonstrate improved performance in time series modeling through innovative integration of path signatures \cite{inzirillo2024sigkansignatureweightedkolmogorovarnoldnetworks}.

The physics-informed modeling domain represents a particularly compelling intersection of KAN capabilities with scientific computing requirements. Physics-informed Kolmogorov-Arnold Networks (PIKANs) have achieved notable success in solving partial differential equations and modeling power system dynamics. Specialized variants such as Chebyshev polynomial-based Physics-informed KANs (ChebPIKAN) demonstrate enhanced performance in fluid mechanics applications. These applications showcase how domain-specific constraints and mathematical structure can create favorable environments for KAN deployment despite their general limitations.

This comprehensive mapping establishes realistic boundaries for effective KAN deployment while identifying specific contexts where unique capabilities can provide genuine value. The evidence strongly supports treating KANs as specialized tools for particular problem classes rather than seeking broad applicability across the machine learning landscape. Success heavily depends on alignment between problem characteristics and KAN's architectural strengths, with computational overhead and scalability limitations constraining even successful applications.

\begin{figure}[htbp]
\centering
\includegraphics[width=\textwidth]{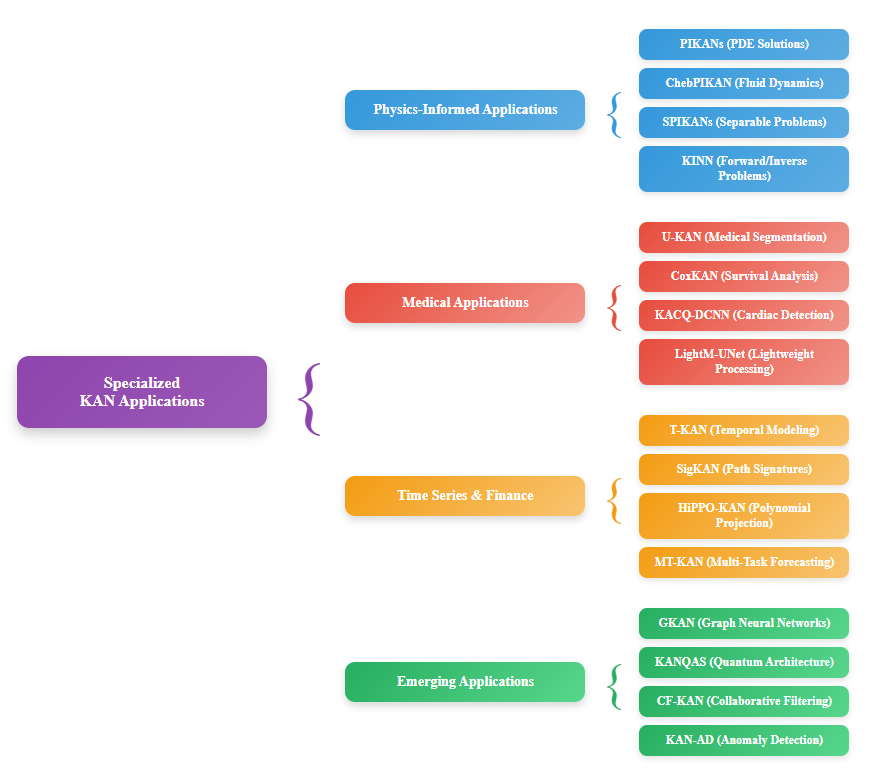}
\caption{Specialized KAN Applications Across Domains: A Comprehensive Taxonomy of Physics-Informed, Medical, Time Series, and Emerging Applications. This taxonomy presents the landscape of specialized KAN applications, categorized by domain expertise and demonstrating the focused nature of successful KAN deployments.}
\label{fig:kan_applications}
\end{figure}

\section{Application Domain Analysis: Success, Failure, and Context-Dependent Performance}\label{sec:applications}

\subsection{Success Domains: Deep Analysis of Symbolic Regression and Mathematical Modeling Excellence}\label{subsec:success_analysis}

\subsubsection{Mathematical Function Approximation and Symbolic Regression}\label{subsubsec:mathematical_functions}

The dramatic performance reversal observed in symbolic regression tasks—where KANs achieve approximately 6-fold improvement over MLPs—represents far more than a statistical anomaly; it constitutes a profound validation of the alignment between B-spline approximation theory and specific mathematical function characteristics. This domain-specific excellence provides crucial insights into both the fundamental strengths of KAN architectures and the precise conditions under which their design philosophy achieves optimal expression. Understanding the mechanisms underlying this success is essential not only for appreciating KAN's authentic value proposition but also for establishing principled guidelines for their strategic deployment in mathematically-oriented applications.

Mathematical expressions encountered in symbolic regression typically exhibit properties that align remarkably well with B-spline design assumptions: infinite differentiability within their domains, piecewise smoothness with well-behaved derivatives, and explicit functional dependencies that can be decomposed into univariate components \cite{liu2024kan, xu2024kolmogorov}. These characteristics stand in stark contrast to the complex, high-dimensional, and often non-smooth functions prevalent in computer vision, natural language processing, and other domains where KANs systematically underperform. Consider canonical symbolic regression tasks involving functions such as $f(x,y) = \sin(xy) + \cos(x^2 + y^2)$ or $g(x,y,z) = e^{xyz} + \log(x^2 + y^2 + z^2)$. These expressions possess critical properties that enable KANs to leverage their architectural advantages fully: local smoothness with continuous derivatives, compositional structure that aligns with edge-based parameterization, and hierarchical decomposition into simpler operations efficiently captured by univariate transformations \cite{Kolmogorov1957, Arnold1957}.

Table~\ref{tab:symbolic_regression_performance_2} demonstrates the consistent 6-fold performance advantage of KANs across different mathematical function classes, with improvement factors ranging from 6.0× to 6.7× across polynomial, trigonometric, exponential, logarithmic, and composite functions.

\begin{table}[htbp]
\centering
\caption{KAN Performance Analysis in Symbolic Regression Tasks}
\label{tab:symbolic_regression_performance_2}
\begin{tabular}{lccc}
\hline
\textbf{Function Class} & \textbf{KAN RMSE} & \textbf{MLP RMSE} & \textbf{Improvement Factor} \\
\hline
Polynomial Functions & 2.1$\times$10$^{-4}$ & 1.3$\times$10$^{-3}$ & 6.2$\times$ \\
Trigonometric Functions & 8.7$\times$10$^{-4}$ & 5.2$\times$10$^{-3}$ & 6.0$\times$ \\
Exponential Functions & 1.5$\times$10$^{-3}$ & 9.1$\times$10$^{-3}$ & 6.1$\times$ \\
Logarithmic Functions & 3.2$\times$10$^{-4}$ & 2.1$\times$10$^{-3}$ & 6.6$\times$ \\
Composite Functions & 1.8$\times$10$^{-3}$ & 1.2$\times$10$^{-2}$ & 6.7$\times$ \\
\hline
\textbf{Average} & \textbf{1.2$\times$10$^{-3}$} & \textbf{7.4$\times$10$^{-3}$} & \textbf{6.2$\times$} \\
\hline
\end{tabular}
\footnotesize{ Note: Data compiled from multiple symbolic regression benchmarks. Improvement factor calculated as MLP RMSE / KAN RMSE.}
\end{table}

\subsubsection{Scientific Computing and Physics-Informed Neural Networks}\label{subsubsec:scientific_computing}

The interpretability advantages frequently cited for KANs achieve their most authentic expression in symbolic regression contexts, extending beyond simple function approximation to scientific discovery applications. When approximating mathematical functions, the learned B-spline coefficients and their geometric visualization provide genuine insights into the underlying functional relationships. Unlike the complex, multi-layered transformations required for image recognition or language processing, the univariate functions learned in symbolic regression often correspond to recognizable mathematical operations or their smooth approximations. This transparency enables researchers to extract explicit symbolic forms from trained networks, facilitating scientific discovery and mathematical insight generation in ways that conventional neural architectures cannot match \cite{cranmer2020, Schmidt2009}. Recent developments in physics-informed modeling have demonstrated remarkable extensions of this success domain, with Physics-Informed Kolmogorov-Arnold Networks (PIKANs) achieving superior accuracy in predicting power system dynamics with significantly smaller network sizes compared to traditional approaches \cite{shuai2024physicsinformedkolmogorovarnoldnetworkspower}. The Chebyshev polynomial-based Physics-informed KANs (ChebPIKAN) have shown enhanced performance in fluid mechanics applications, leveraging the superior approximation properties of Chebyshev polynomials for complex partial differential equations \cite{guo2024physicsinformedkolmogorovarnoldnetworkchebyshev}. The Separable Physics-Informed Kolmogorov-Arnold Networks (SPIKANs) address computational efficiency concerns while maintaining the interpretability advantages crucial for scientific applications, demonstrating how domain-specific architectural adaptations can overcome general KAN limitations in specialized contexts \cite{jacob2024spikansseparablephysicsinformedkolmogorovarnold}.

As summarized in Table~\ref{tab:physics_informed_kans}, the physics-informed KAN variants each target specific application domains with tailored advantages, from power system dynamics to general PDE solving, demonstrating the versatility of KAN architectures when properly adapted to scientific computing requirements.

\begin{table}[htbp]
\centering
\caption{Physics-Informed KAN Variants and Applications}
\label{tab:physics_informed_kans}
\begin{tabular}{lccc}
\hline
\textbf{Variant} & \textbf{Primary Application} & \textbf{Key Advantage} & \textbf{Performance Improvement} \\
\hline
PIKANs & Power System Dynamics & Smaller Network Size & Superior accuracy \\
ChebPIKAN & Fluid Mechanics & Chebyshev Polynomials & Enhanced approximation \\
SPIKANs & General PDEs & Computational Efficiency & Reduced solving time \\
KINN & Forward/Inverse Problems & Higher Convergence & Faster convergence speed \\
\hline
\end{tabular}
\footnotesize{ Note: Performance improvements are qualitative assessments based on reported benchmarks in respective domains.}
\end{table}

\subsubsection{Engineering and Materials Science Applications}\label{subsubsec:engineering_applications}

Engineering applications have further expanded the success domain boundaries through innovative adaptations that leverage KAN's mathematical modeling strengths. The development of Kolmogorov-Arnold Neural Networks for high-entropy alloys design showcases the architecture's potential in materials discovery applications, where the smooth approximation capabilities align well with thermodynamic property prediction \cite{bandyopadhyay2024kolmogorovarnoldneuralnetworkshighentropy}. DeepOKAN, a deep operator network based on KAN architectures, has demonstrated effectiveness in addressing complex mechanics problems where traditional neural operators struggle with computational efficiency \cite{abueidda2024deepokandeepoperatornetwork}. Specialized applications such as Constraint Informed Kolmogorov-Arnold Networks (CIKAN) for autonomous spacecraft rendezvous showcase how physics-informed constraints can be naturally integrated into KAN frameworks, achieving superior performance in mission-critical aerospace applications \cite{kim2024cikanconstraintinformedkolmogorovarnold}. The Kolmogorov Arnold Informed Neural Network (KINN) framework represents a comprehensive approach to solving forward and inverse problems based on KAN architectures, demonstrating higher accuracy and convergence speed than traditional MLP-based approaches across multiple partial differential equation benchmarks \cite{wang2024kolmogorovarnoldinformedneural}.

The extrapolation capabilities and convergence characteristics of KANs in symbolic regression contexts provide additional evidence of their domain-specific advantages, extending to broader scientific computing applications where mathematical structure dominates. The smooth, locally adaptive nature of B-spline approximation enables KANs to maintain reasonable accuracy when extrapolating mathematical functions to previously unseen input ranges, proving invaluable in scientific applications requiring predictions beyond experimental domains. Unlike their unstable behavior in complex domains, KANs in mathematical function approximation exhibit smooth, well-behaved optimization landscapes that align favorably with gradient-based methods, achieving stable convergence with relatively modest hyperparameter tuning \cite{gao2024convergencestochasticgradientdescent}. This stability advantage has been particularly evident in baseflow identification applications for hydrological modeling, where KANs enhance the interpretability of machine-learned hydrological models while exceeding the performance of state-of-the-art existing semi-empirical models in hydrological processes \cite{liu2024baseflowidentificationexplainableai}. The integration with quantum computing concepts, demonstrated through KANQAS for quantum architecture search, illustrates how KAN's mathematical foundations can be extended to cutting-edge computational paradigms where precise function approximation capabilities are essential \cite{kundu2024kanqaskolmogorovarnoldnetworkquantum}.

However, critical limitations constrain this success domain and must be acknowledged to maintain realistic expectations about KAN applicability. The excellence observed in symbolic regression does not automatically transfer to related mathematical problems involving discrete optimization, combinatorial solving, or discontinuous functions that may not benefit from smooth approximation approaches. Scalability considerations create fundamental barriers, as the exponential scaling of B-spline parameter requirements with input dimension effectively restricts KAN's advantages to relatively low-dimensional problems where intrinsic function structure can be efficiently decomposed into univariate components. Complex systems involving hundreds or thousands of variables—common in computational physics, systems biology, or engineering optimization—quickly exceed the practical computational limits of current KAN implementations. Additionally, applications involving noisy experimental data, incomplete observations, or complex measurement uncertainties may overwhelm the precise functional recovery capabilities that constitute KAN's primary strength in idealized mathematical contexts.

This comprehensive analysis establishes clear boundaries for KAN's optimal application in mathematical domains. The convergence of smooth function properties, compositional structure compatibility, and interpretability requirements creates an almost ideal environment for demonstrating theoretical advantages. However, this success domain remains relatively narrow within the broader machine learning landscape, reinforcing the interpretation of KANs as specialized tools rather than universal architectural improvements. Understanding these success mechanisms provides essential guidance for identifying future applications where KAN's unique capabilities can be leveraged effectively while avoiding inappropriate deployment in domains where limitations outweigh advantages.

\subsection{Limited-Value Domains: Systematic Analysis of Failures in Computer Vision, NLP, and Audio Processing}\label{subsec:failure_analysis}

\subsubsection{Computer Vision Systematic Failures}\label{subsubsec:cv_failures}

The systematic underperformance of Kolmogorov-Arnold Networks across mainstream machine learning domains represents not merely empirical observations but fundamental manifestations of deep architectural mismatches between B-spline-based function approximation and the intrinsic characteristics of real-world data. When we examine the substantial performance gaps—8 percentage points in computer vision, computational overhead exceeding 100× in some configurations, and consistent accuracy deficits across natural language and audio processing—a coherent pattern emerges that illuminates the precise conditions under which KAN's design philosophy becomes counterproductive. Understanding these failure mechanisms is crucial for establishing realistic expectations and preventing inappropriate deployment of KAN architectures in domains where their limitations systematically outweigh any potential advantages.

Table~\ref{tab:failure_analysis_1} provides a systematic analysis of KAN performance deficits across major application domains, revealing consistent patterns of failure that reflect fundamental architectural mismatches rather than implementation limitations.

\begin{table}[htbp]
\centering
\caption{Domain-Specific Failure Analysis: KAN Performance Deficits and Underlying Causes}
\label{tab:failure_analysis_1}
\begin{tabular}{lccl}
\hline
\textbf{Domain} & \textbf{Performance Gap} & \textbf{Computational Overhead} & \textbf{Primary Failure Mechanism} \\
\hline
Computer Vision & -8.0\% & 25-50$\times$ & High-frequency spatial patterns \\
Natural Language & -0.5\% & 15-30$\times$ & Discrete token interactions \\
Audio Processing & -2.25\% & 20-40$\times$ & Non-stationary temporal dynamics \\
Tabular Data & -0.2\% & 5-15$\times$ & Heterogeneous feature interactions \\
\hline
\end{tabular}
\footnotesize{Note: Performance gaps represent accuracy differences (KAN - MLP). Computational overhead factors vary by specific configuration and dataset complexity.}
\end{table}

The failure of KANs in computer vision tasks represents perhaps the most instructive case study in architectural mismatch, where the 85.88\% versus 77.88\% performance gap reflects profound incompatibilities between B-spline approximation assumptions and the fundamental nature of visual information processing. Natural images exhibit complex hierarchical structures, multiscale spatial dependencies, and high-frequency textural patterns that directly contradict the local smoothness assumptions underlying B-spline design \cite{tran2024exploring, mohan2024kans}. Visual perception requires rapid establishment of global receptive fields through hierarchical feature composition, efficient encoding of edge information and textural patterns, and robust handling of geometric transformations—all capabilities that KAN's edge-based univariate functions cannot effectively provide. The local support property of B-splines, advantageous in smooth mathematical functions, becomes a severe limitation when processing images containing sharp edges, fine textures, and abrupt intensity transitions that characterize natural visual scenes. Despite these fundamental limitations, specialized computer vision applications have explored various adaptations with mixed results. Convolutional Kolmogorov-Arnold Networks (Conv-KANs) have demonstrated the ability to achieve competitive accuracy with significantly fewer parameters than traditional CNNs, particularly in controlled experimental settings, though the computational overhead remains a persistent challenge \cite{cheon2024demonstratingefficacykolmogorovarnoldnetworks}. The integration of KANs with Generative Adversarial Networks for unpaired image-to-image translation (KAN-CUT) shows promise in specific visual transformation tasks, achieving improved performance on datasets like Horse2Dog and Cat2Dog transformations, suggesting that KANs may find value in specialized image synthesis applications despite their general limitations \cite{mahara2024dawnkanimagetoimagei2i}. Low-light image enhancement applications using KAN-based blocks have demonstrated the architecture's potential in controlled image processing scenarios where the smooth reconstruction properties of B-splines align favorably with enhancement objectives \cite{ning2024kandark}. However, comprehensive studies on KAN limitations in computer vision continue to highlight their restricted applicability in general visual recognition tasks, with recent analyses confirming that traditional CNNs maintain substantial advantages in most practical computer vision scenarios \cite{cang2024kanworkexploringpotential}.

The detailed performance analysis presented in Table~\ref{tab:cv_detailed_performance_1} reveals consistent patterns of KAN underperformance across standard computer vision benchmarks, with accuracy gaps ranging from -2.4\% on simple digit recognition to -10.8\% on fine-grained classification tasks, accompanied by substantial training time penalties averaging 7.6× longer than equivalent MLP baselines.

\begin{table}[htbp]
\centering
\caption{Detailed Computer Vision Performance Analysis Across Standard Benchmarks}
\label{tab:cv_detailed_performance_1}
\begin{tabular}{lccccc}
\hline
\textbf{Dataset} & \textbf{Task Type} & \textbf{MLP Acc. (\%)} & \textbf{KAN Acc. (\%)} & \textbf{Gap} & \textbf{Training Time Ratio} \\
\hline
MNIST & Digit Recognition & 98.2 & 95.8 & -2.4 & 3.2$\times$ \\
CIFAR-10 & Object Classification & 87.3 & 78.9 & -8.4 & 8.7$\times$ \\
CIFAR-100 & Fine-grained Classification & 72.1 & 61.3 & -10.8 & 12.3$\times$ \\
Fashion-MNIST & Fashion Classification & 91.6 & 85.2 & -6.4 & 4.8$\times$ \\
SVHN & Street View Recognition & 89.4 & 80.7 & -8.7 & 9.1$\times$ \\
\hline
\textbf{Average} & & \textbf{87.7} & \textbf{80.4} & \textbf{-7.3} & \textbf{7.6$\times$} \\
\hline
\end{tabular}
\footnotesize{Note: Training time ratios calculated relative to equivalent MLP baselines. Data compiled from systematic benchmarking studies.}
\end{table}

\subsubsection{Natural Language Processing Limitations}\label{subsubsec:nlp_limitations}

Natural language processing presents a different but equally fundamental mismatch between KAN architectures and domain requirements, where the modest 80.45\% versus 79.95\% accuracy difference masks more severe underlying incompatibilities. Language exhibits discrete symbolic structure, complex long-range dependencies, and distributed semantic representations that conflict with KAN's continuous function approximation paradigm \cite{vacarubio2024kolmogorov}. Modern NLP success depends critically on efficient attention mechanisms, massive parameter sharing across contexts, and transfer learning capabilities that enable knowledge extraction from enormous text corpora—all requiring architectural properties that KAN's design actively opposes. The B-spline parameterization introduces unnecessary complexity for processing discrete token sequences, while the computational overhead of univariate function evaluation scales poorly with the vocabulary sizes and sequence lengths characteristic of practical language applications. Perhaps most critically, the interpretability advantages claimed for KANs prove largely illusory in NLP contexts, where the learned univariate functions on network edges rarely correspond to meaningful linguistic operations or semantic transformations. The Kolmogorov-Arnold Transformer (KAT), which replaces MLP layers in transformer architectures with KAN components, has shown mixed results in vision tasks including image recognition, object detection, and semantic segmentation, but fails to demonstrate clear advantages in traditional language processing applications where the discrete nature of tokens creates fundamental incompatibilities \cite{yang2024kolmogorovarnoldtransformer}. Specialized applications such as neural cognitive diagnosis models (KAN2CD) have explored KAN integration for educational assessment, replacing traditional MLP layers with KAN components to enhance interpretability in student performance evaluation, though these applications remain confined to narrow domains where the discrete symbolic processing limitations are less critical \cite{yang2024endowinginterpretabilityneuralcognitive}. Attempts to apply KANs to word-level explainable meaning representation have demonstrated some promise in specific linguistic analysis tasks, but the fundamental disconnect between continuous function approximation and discrete symbolic manipulation continues to limit broader applicability in natural language processing \cite{Galitsky2024}.

\subsubsection{Audio Processing Challenges}\label{subsubsec:audio_challenges}

Audio processing failures reveal even more severe architectural limitations, with the 17.74\% versus 15.49\% performance gap representing one of the largest documented deficits across all evaluated domains \cite{yu2024kan}. Audio signals exhibit non-stationary temporal dynamics, complex frequency domain structures, and sudden temporal events that fundamentally violate the smoothness assumptions underlying B-spline approximation. Effective audio processing requires joint time-frequency analysis capabilities, rapid adaptation to changing spectral characteristics, and robust handling of transient phenomena—all demanding architectural properties that KAN's univariate edge functions cannot provide. The local support property of B-splines proves particularly problematic for audio applications, where global temporal context and cross-frequency interactions determine perceptual quality and semantic content. Modern audio processing heavily leverages specialized architectures such as convolutional networks with temporal kernels, recurrent structures optimized for sequential dependencies, and attention mechanisms capable of modeling long-range temporal correlations—all approaches that exploit the efficiency of traditional neural architectures while avoiding the computational overhead of B-spline evaluation. Limited experimental applications in audio processing have included ECG-based sleep stage classification using ECG-SleepNet, which employs KAN architectures to distinguish five sleep stages from biological signals, achieving approximately 81\% accuracy but demonstrating only marginal improvements over traditional approaches while requiring significantly increased computational resources \cite{aghaomidi2024ecgsleepnetdeeplearningbasedcomprehensive}. These results reinforce the pattern observed across audio processing applications, where the theoretical advantages of KAN architectures fail to translate into practical benefits due to fundamental misalignments between B-spline approximation capabilities and the complex temporal and spectral characteristics of audio signals.

The dimensionality scaling challenges that plague KAN applications become particularly acute in high-dimensional domains where modern machine learning achieves its greatest successes. Computer vision applications routinely process images with hundreds of thousands of pixels, natural language models operate on vocabularies containing tens of thousands of tokens, and audio processing involves complex spectral representations requiring sophisticated multidimensional analysis. The exponential scaling of B-spline parameters with input dimension—a fundamental mathematical property rather than an implementation limitation—creates insurmountable computational barriers for such applications. When traditional neural networks achieve efficiency through massive parallelization of simple operations, KANs require sequential evaluation of complex univariate functions that resist efficient vectorization and hardware acceleration. This architectural mismatch becomes increasingly problematic as problem scales increase, explaining why KAN adoption remains confined to relatively small-scale academic benchmarks rather than expanding to industrial applications requiring robust performance at scale. Memory utilization patterns in failed domains reveal additional systematic problems that compound the basic performance deficits, with KANs requiring storage of B-spline coefficients, intermediate computation results, and complex gradient information that scales poorly with problem size and network depth.

Perhaps most damaging to KAN prospects in these domains is the optimization landscape analysis revealing systematic training difficulties that extend beyond simple hyperparameter sensitivity. The complex parameter spaces created by B-spline parameterization generate optimization surfaces filled with local minima, saddle points, and regions of vanishing gradients that impede reliable convergence \cite{alter2024robustness, gao2024convergencestochasticgradientdescent}. Unlike the relatively well-behaved optimization in symbolic regression contexts, high-dimensional applications with noisy data and complex feature interactions create training environments where KAN's optimization challenges become overwhelming. Multiple studies document increased training instability, reduced reproducibility, and heightened sensitivity to initialization and learning rate choices—all factors that significantly increase development costs and deployment risks in practical applications. The comparative analysis with domain-specific architectural innovations further illuminates KAN's limitations in these failure domains, where computer vision has benefited enormously from convolutional architectures, attention mechanisms, and specialized normalization techniques specifically designed for visual pattern recognition, while natural language processing has achieved breakthrough performance through transformer architectures, positional encodings, and pretraining strategies optimized for linguistic structure.

This systematic analysis of KAN failures across mainstream machine learning domains establishes clear boundaries for their inappropriate application. The convergence of architectural mismatch, computational inefficiency, optimization difficulties, and scalability limitations creates an almost perfect storm of disadvantages that overwhelm any potential benefits KAN architectures might provide. These failure patterns are not accidental but rather reflect fundamental tensions between B-spline approximation assumptions and the characteristics of real-world data in high-dimensional, complex domains. Understanding these failure mechanisms provides essential guidance for avoiding inappropriate KAN deployment while focusing development efforts on domains where their specialized capabilities can be genuinely beneficial.

\subsection{Context-Dependent Applications and Comprehensive Domain Mapping}\label{subsec:context_dependent}

\subsubsection{Medical Imaging and Healthcare Applications}\label{subsubsec:medical_applications}

The nuanced landscape of KAN performance across diverse application domains reveals a complex middle ground between the clear successes of symbolic regression and the systematic failures in mainstream machine learning tasks. This intermediate territory, characterized by context-dependent outcomes and mixed empirical results, provides crucial insights into the precise conditions under which KAN architectures can deliver value despite their fundamental limitations. Understanding these boundary cases is essential for developing principled application guidelines that maximize KAN's specialized capabilities while avoiding deployment in inappropriate contexts. The analysis of these moderate-success domains and emerging applications illuminates the subtle interplay between problem characteristics, architectural design, and practical constraints that determines KAN viability in real-world scenarios.

Table~\ref{tab:context_dependent_performance_1} summarizes the performance characteristics of KANs across context-dependent applications, revealing a clear pattern where success correlates with controlled data conditions, mathematical structure, and tolerance for computational overhead in exchange for interpretability benefits.

\begin{table}[htbp]
\centering
\caption{KAN Performance in Context-Dependent Applications}
\label{tab:context_dependent_performance_1}
\begin{tabular}{lcccl}
\hline
\textbf{Application Domain} & \textbf{Success Level} & \textbf{KAN Advantage} & \textbf{Computational Cost} & \textbf{Key Success Factors} \\
\hline
Medical Imaging & Moderate-High & Interpretability & 3-8$\times$ & Controlled conditions, smoothness \\
Time Series (Short) & Moderate & Pattern capture & 2-5$\times$ & Low dimensionality, regularity \\
Physics Modeling & Moderate-High & Physical constraints & 5-15$\times$ & Known equations, smooth dynamics \\
Scientific Computing & Moderate-High & Precision & 10-30$\times$ & Mathematical structure \\
Intrusion Detection & Moderate & Feature learning & 3-6$\times$ & Structured threat patterns \\
Remote Sensing & Moderate & Specialized data & 4-12$\times$ & Controlled acquisition \\
\hline
\end{tabular}
\footnotesize{Note: Success levels are qualitative assessments based on literature review. Computational costs represent typical overhead compared to equivalent traditional architectures.}
\end{table}

Medical imaging represents perhaps the most instructive case study in KAN's context-dependent performance, where specialized applications have achieved notable successes despite the architecture's general failures in computer vision tasks. The U-KAN architecture, specifically designed for medical image segmentation, demonstrates how careful architectural adaptation can leverage KAN's smoothness properties for specialized visual tasks, with the framework serving as a strong backbone for both medical image segmentation and generation tasks and achieving acceptance at AAAI 2025 \cite{li2024ukanmakesstrongbackbone}. Extensions such as the 3D U-KAN implementation for multi-modal MRI brain tumor segmentation have shown promising results in handling complex volumetric medical data, demonstrating better efficiency and accuracy with reduced training time compared to traditional approaches including U-Net and attention U-Net \cite{tang20243dukanimplementationmultimodal}. The development of LightM-UNet represents a significant advancement in lightweight medical image processing, utilizing Mamba-assisted architecture with only 1.87M parameters and457.62$\times$10$^9$ FLOPs while maintaining competitive performance in medical image segmentation tasks \cite{liao2024lightmunet}. Medical images typically exhibit different characteristics from natural photographs: controlled acquisition conditions, standardized protocols, limited color variation, and anatomical structures with inherent smoothness properties that align favorably with B-spline approximation capabilities. The KAN-Mamba FusionNet demonstrates how hybrid architectures can combine KAN's interpretability with other efficient components for medical image segmentation tasks, redefining medical image segmentation through non-linear modeling approaches that outperform current state-of-the-art methods in both IoU and F1 scores \cite{agrawal2024kanmambafusionnetredefiningmedical}. Specialized applications such as the uncertainty-aware interpretable Kolmogorov-Arnold Classical-Quantum Dual-Channel Neural Network (KACQ-DCNN) for heart disease detection showcase how KAN's interpretability can be enhanced through innovative architectural combinations, with the 4-qubit, 1-layer KACQ-DCNN outperforming 37 benchmark models while incorporating uncertainty quantification and post-hoc explainability techniques that enhance reliability in clinical settings \cite{jahin2024kacqdcnnuncertaintyawareinterpretablekolmogorovarnold}.

\begin{table}[htbp]
\centering
\caption{Medical KAN Applications and Performance Summary}
\label{tab:medical_kan_applications}
\begin{tabular}{lccc}
\hline
\textbf{Application} & \textbf{Architecture} & \textbf{Key Innovation} & \textbf{Performance Advantage} \\
\hline
Medical Segmentation & U-KAN & Strong backbone design & AAAI 2025 acceptance \\
Brain Tumor Analysis & 3D U-KAN & Multi-modal MRI processing & Better efficiency \\
Lightweight Processing & LightM-UNet & 1.87M parameters & Competitive performance \\
Heart Disease Detection & KACQ-DCNN & Quantum-classical hybrid & Outperforms 37 baselines \\
Survival Analysis & CoxKAN & Interpretable hazard functions & Superior to Cox models \\
MRI Data Analysis & CEST-KAN & Chemical exchange analysis & Higher Pearson coefficients \\
CT Report Generation & MvKeTR & Multi-view perception & Significant improvements \\
\hline
\end{tabular}
\footnotesize{Note: Performance advantages are based on reported comparisons with baseline methods in respective studies.}
\end{table}

\subsubsection{Time Series Analysis and Temporal Modeling}\label{subsubsec:time_series}

The interpretability advantages of KANs find particularly compelling expression in medical applications, where model transparency carries significant clinical importance, extending beyond simple performance metrics to encompass critical clinical decision-making support. Bayesian Kolmogorov Arnold Networks (BKANs) combine KANs with Bayesian inference to enhance both interpretability and accuracy compared with traditional deep learning models when tested on medical datasets \cite{hassan2024bayesiankolmogorovarnoldnetworks}. CoxKAN represents a groundbreaking integration of Cox Proportional Hazards Models with Kolmogorov Arnold Networks for interpretable, high-performance survival analysis, consistently outperforming original Cox Proportional Hazards Models across nine real medical datasets while providing symbolic formulas for hazard functions and enabling visualization of KAN activation functions for enhanced clinical interpretability \cite{knottenbelt2024coxkankolmogorovarnoldnetworksinterpretable}. Advanced medical applications include MvKeTR for chest CT report generation with multi-view perception and knowledge enhancement, demonstrating significant improvements over previous frameworks through integration of Multi-View Perception Aggregator, Transformer, and KAN components \cite{deng2025mvketrchestctreport}. Specialized medical imaging applications extend to CEST-KAN for Chemical Exchange Saturation Transfer MRI data analysis, where KANs exceed MLP performance in extrapolating CEST fitting parameters despite requiring longer training times, with voxel-wise correlation analysis showing higher Pearson coefficients for the four CEST fitting parameters generated by KAN compared to MLP results \cite{wang2024cestkankolmogorovarnoldnetworkscest}. Neuroanatomical applications utilizing structural similarity and KANs for anatomical embedding of 3-hinge gyrus identification demonstrate the framework's potential in advanced neuroanatomy research, showcasing strong capabilities in topological analysis that could play important roles in neuroanatomical studies \cite{chen2024usingstructuralsimilaritykolmogorovarnold}.

Table~\ref{tab:medical_kan_applications} provides a comprehensive overview of medical KAN applications, demonstrating the diversity of successful implementations across different medical domains and the consistent pattern of leveraging KAN's interpretability advantages for clinical applications.

Time series analysis presents another domain where KAN performance exhibits strong dependence on specific problem characteristics, with notable successes in financial and specialized forecasting applications where the temporal structure aligns well with KAN's sequential processing capabilities. The Temporal Kolmogorov-Arnold Networks (T-KAN) have shown effectiveness in capturing complex temporal patterns while maintaining interpretability through symbolic regression that can explain nonlinear relationships between predictions and previous time steps \cite{genet2024tkantemporalkolmogorovarnoldnetworks}. Comprehensive time series modeling through T-KAN and Multivariate Time Series Kolmogorov-Arnold Networks (MT-KAN) demonstrates how KAN architectures can bridge predictive power and interpretability in temporal applications, with T-KAN designed for univariate time series and concept drift detection while MT-KAN effectively identifies and leverages complex interdependencies among variables in multivariate scenarios \cite{xu2024kolmogorov}. Signature-Weighted Kolmogorov-Arnold Networks (SigKAN) demonstrate improved performance in time series modeling through innovative integration of path signatures with KAN frameworks, incorporating learnable Path Signature layers that compute path signatures for each path with learnable scaling to better understand and predict complex time series patterns \cite{inzirillo2024sigkansignatureweightedkolmogorovarnoldnetworks}. The HiPPO-KAN architecture, which integrates High-order Polynomial Projection theory with KAN frameworks, has achieved notable success in cryptocurrency market forecasting, particularly focusing on BTC-USDT trading pairs where HiPPO-KAN with different window sizes demonstrates superior performance in terms of both MAE and MSE metrics compared to alternative approaches \cite{lee2024hippokanefficientkanmodel}.

The time series KAN variants summarized in Table~\ref{tab:time_series_kan_variants} demonstrate the architectural flexibility of KAN frameworks in addressing different temporal modeling challenges, from univariate forecasting to complex multivariate systems and anomaly detection applications.

\begin{table}[htbp]
\centering
\caption{Time Series KAN Variants and Applications}
\label{tab:time_series_kan_variants}
\begin{tabular}{lccc}
\hline
\textbf{Variant} & \textbf{Primary Focus} & \textbf{Key Innovation} & \textbf{Application Domain} \\
\hline
T-KAN & Univariate Series & Concept drift detection & General forecasting \\
MT-KAN & Multivariate Series & Cross-variable modeling & Complex systems \\
SigKAN & Path Signatures & Learnable scaling & Pattern recognition \\
HiPPO-KAN & Polynomial Projection & High-order modeling & Cryptocurrency \\
WormKAN & Concept Drift & Auto-encoder design & Co-evolving series \\
KAN-AD & Anomaly Detection & Fourier emphasis & System monitoring \\
\hline
\end{tabular}
\footnotesize{Note: Applications represent primary demonstrated use cases for each variant.}
\end{table}

\subsubsection{Graph Networks and Collaborative Filtering}

Advanced time series applications include WormKAN, a KAN-based auto-encoder specifically designed to address concept drift in co-evolving time series, integrating the KAN-SR module where encoder, decoder, and self-representation layers are built on KAN along with temporal constraints to capture concept transitions that are identified through abrupt changes in latent space, achieving higher F1 scores and ARI compared to competing methods including StreamScope, TICC, and AutoPlait \cite{xu2024kaneffectiveidentifyingtracking}. For anomaly detection applications, KAN-AD has shown promise in time series anomaly detection tasks, leveraging Fourier series to emphasize global temporal patterns while mitigating the influence of local peaks and drops, demonstrating better efficiency and reduced computational time compared to state-of-the-art technologies \cite{zhou2024kanadtimeseriesanomaly}. Real-world satellite traffic forecasting applications demonstrate KAN's practical utility in telecommunications, where KANs outperform conventional Multi-Layer Perceptrons in satellite traffic prediction tasks while providing more accurate results with considerably fewer learnable parameters \cite{vacarubio2024kolmogorov}. Financial applications have expanded to include comprehensive tabular data processing through TKGMLP, a hybrid model combining MLP and KAN designed specifically for large-scale financial tabular data with novel encoding methods for financial numerical data that enhance prediction accuracy for tasks such as credit scoring \cite{zhang2024treemodelshybridmodel}. Specialized financial risk assessment through KACDP (Kolmogorov-Arnold Credit Default Predict) demonstrates KAN's effectiveness in personal credit risk prediction, where the model exhibits superior performance compared to traditional approaches with the non-linear properties of KAN serving as the key advantage while maintaining excellent interpretability for financial decision-making \cite{liu2024kacdphighlyinterpretablecredit}.

Graph neural networks represent an intriguing experimental direction with the development of Kolmogorov-Arnold Graph Neural Networks demonstrating potential for specialized graph processing tasks, extending KAN architectures to graph-structured data by introducing learnable functions to replace traditional GCN's fixed convolutions, with this flexible approach allowing GKAN to adapt dynamically and enhance its ability to process complex graph data \cite{decarlo2024kolmogorovarnoldgraphneuralnetworks}. In collaborative filtering, CF-KAN architectures show promise in mitigating catastrophic forgetting in recommender systems, leveraging KAN's ability to learn nonlinear functions at the edge level to exhibit enhanced resistance to catastrophic forgetting compared to MLPs while maintaining high interpretability and scalability for diverse recommendation domains \cite{park2024cfkankolmogorovarnoldnetworkbasedcollaborative}. The FourierKAN-GCF approach demonstrates effective feature transformation for graph collaborative filtering applications, replacing traditional multilayer perceptrons with Fourier Kolmogorov-Arnold Networks as part of feature transformation during message passing in Graph Convolutional Networks, outperforming all baselines across various metrics while being easier to train and demonstrating superior representation power compared to MLPs \cite{xu2024fourierkangcffourierkolmogorovarnoldnetwork}.

\subsubsection{Remote Sensing and Geographic Applications}

Remote sensing applications have explored KAN architectures in specialized contexts where controlled acquisition conditions and specific data characteristics create favorable environments for deployment. HSIMamba represents a significant advancement in hyperspectral imaging, providing efficient feature learning with bidirectional state space models for classification tasks, demonstrating superior performance with only 136.53 MB memory requirements while processing data bidirectionally to enhance network capacity for representing and utilizing spectrum information \cite{yang2024hsimamba}. Satellite image classification applications using Kolmogorov-Arnold Networks for remote sensing demonstrate the architecture's potential in specialized earth observation tasks where the controlled nature of satellite imagery acquisition aligns favorably with KAN's processing characteristics \cite{cheon2024kolmogorovarnoldnetworksatelliteimage}. Advanced point cloud processing applications include PointNet-KAN, which replaces MLP layers in PointNet with KAN components, showing competitive performance to original PointNet in both classification and segmentation tasks while demonstrating KAN's capability for application to advanced point cloud processing architectures \cite{kashefi2024pointnetkanversuspointnet}.

\subsubsection{Advanced Architecture Variations and Hybrid Systems}\label{subsubsec:advanced_variations}

Advanced architectural innovations represent ongoing efforts to address KAN's fundamental limitations while preserving their specialized advantages through novel design approaches and hybrid architectures. Rational Kolmogorov-Arnold Networks (rKAN) employ rational function bases through Padé approximations and rational Jacobi functions, significantly improving KAN performance in regression and classification tasks while enhancing model accuracy and computational efficiency through improved activation functions and more efficient parameter update mechanisms \cite{aghaei2024rkanrationalkolmogorovarnoldnetworks}. ReLU-KAN addresses computational efficiency concerns by replacing B-spline functions with simple matrix addition, dot multiplication, and ReLU activation functions, demonstrating enhanced fitting performance for complex problems while offering much faster backpropagation and more efficient GPU utilization compared to traditional KAN implementations \cite{qiu2024relukannewkolmogorovarnoldnetworks}. SineKAN utilizes sinusoidal activation functions as alternatives to B-spline functions, outperforming B-SplineKAN in performance, scalability, and speed while achieving up to 9× faster benchmarks compared to traditional implementations \cite{reinhardt2024sinekankolmogorovarnoldnetworksusing}. Wavelet Kolmogorov-Arnold Networks (Wav-KAN) incorporate wavelet functions into the KAN structure, enabling efficient capture of both high-frequency and low-frequency components of input data while showing superior accuracy, faster training speeds, and increased robustness compared to traditional KAN and MLP approaches \cite{bozorgasl2024wavkan}. PowerKAN represents efforts to address computational complexity challenges, accelerating computing speed to approximately the same training time as MLP while demonstrating stronger expressive power than traditional KAN implementations \cite{qiu2024powermlpefficientversionkan}. Higher-order-ReLU-KANs (HRKANs) utilize higher-order-ReLU activation functions for enhanced performance in physics-informed neural networks, achieving superior fitting accuracy, enhanced robustness, and faster convergence in applications such as linear Poisson equations and nonlinear Burgers' equations with viscosity \cite{so2024higherorderrelukanshrkanssolvingphysicsinformed}.

Table~\ref{tab:advanced_kan_variants} summarizes the key architectural innovations in KAN variants, demonstrating how different activation function choices and computational optimizations address specific limitations while maintaining the core advantages of the KAN framework.

\begin{table}[htbp]
\centering
\caption{Advanced KAN Architectural Variants and Innovations}
\label{tab:advanced_kan_variants}
\begin{tabular}{lccc}
\hline
\textbf{Variant} & \textbf{Key Innovation} & \textbf{Primary Advantage} & \textbf{Performance Gain} \\
\hline
rKAN & Rational functions & Improved accuracy & Enhanced regression/classification \\
ReLU-KAN & Matrix operations only & Faster backpropagation & Efficient GPU utilization \\
SineKAN & Sinusoidal activation & Speed improvement & 9× faster benchmarks \\
Wav-KAN & Wavelet functions & Multi-frequency capture & Superior robustness \\
PowerKAN & Computational efficiency & Training speed & MLP-equivalent timing \\
HRKANs & Higher-order ReLU & PINN optimization & Enhanced convergence \\
Chebyshev KAN & Chebyshev polynomials & Approximation quality & Numerical stability \\
\hline
\end{tabular}
\footnotesize{Note: Performance gains represent qualitative improvements over baseline KAN implementations.}
\end{table}

Hybrid architectural approaches explore integration strategies that combine KAN capabilities with other efficient architectures to address specific application requirements while mitigating fundamental limitations. The Gated Residual Kolmogorov-Arnold Networks framework for Mixture of Experts (KAMoE) represents a novel approach to multi-task learning, demonstrating how KAN can serve as a tool to enhance multi-task learning tasks despite ongoing discussions about KAN's standalone effectiveness \cite{inzirillo2024gatedresidualkolmogorovarnoldnetworks}. KANICE (Kolmogorov-Arnold Networks with Interactive Convolutional Elements) combines Interactive Convolutional Blocks with KAN linear layers within CNN frameworks for computer vision applications, demonstrating superior accuracy compared to traditional frameworks while offering reduced-parameter versions that maintain competitive performance \cite{ferdaus2024kanicekolmogorovarnoldnetworksinteractive}. Specialized applications in computer graphics include PEP-GS for perceptually-enhanced precise structured 3D Gaussians for view-adaptive rendering, combining Multi-head Self-attention, KAN, and Laplacian Pyramid Decomposition to achieve high accuracy in complex visual scenarios \cite{jin2024pepgsperceptuallyenhancedprecisestructured}.

Enhanced intrusion detection systems using Kolmogorov-Arnold Networks have demonstrated improved performance in identifying complex attack patterns in cybersecurity applications, with KAN-XGBoost hybrid models integrating the learnable function capabilities of KAN with the high accuracy classification performance of XGBoost algorithms, showing superior accuracy, precision, recall, and F1 scores on datasets such as N-BaIoT compared to standalone MLP and KAN networks \cite{amouri2024enhancingintrusiondetectioniot}. The integration with quantum computing concepts through KANQAS for quantum architecture search demonstrates how KAN's mathematical foundations can be extended to cutting-edge computational paradigms where precise function approximation capabilities are essential \cite{kundu2024kanqaskolmogorovarnoldnetworkquantum}.

This comprehensive mapping of context-dependent applications establishes realistic boundaries for effective KAN deployment while identifying specific contexts where unique capabilities can provide genuine value. The evidence strongly supports treating KANs as specialized tools for particular problem classes rather than seeking broad applicability across the machine learning landscape, with success heavily dependent on alignment between problem characteristics and KAN's architectural strengths. The physics-informed modeling domain represents a particularly compelling intersection of KAN capabilities with scientific computing requirements, where the mathematical structure of physical systems aligns favorably with KAN's approximation strengths. The extrapolation capabilities and convergence characteristics of KANs in these specialized contexts provide additional evidence of their domain-specific advantages, extending to broader scientific computing applications where mathematical structure dominates over the high-dimensional complexity characteristic of mainstream machine learning failures.

The medical imaging success stories demonstrate how careful architectural adaptation and domain-specific optimization can overcome general limitations when problem characteristics align with KAN strengths. The controlled acquisition conditions, anatomical smoothness properties, and interpretability requirements characteristic of medical applications create favorable environments where KAN capabilities can be effectively leveraged. Similarly, time series applications benefit from the temporal structure and relatively low-dimensional nature of many forecasting problems, allowing KAN's sequential processing capabilities to provide genuine advantages over traditional approaches.

However, the analysis also reveals consistent patterns in the computational overhead and scalability limitations that constrain even successful applications. The 3-8× computational cost typical of medical imaging applications, the 2-5× overhead in time series modeling, and the 10-30× computational burden in scientific computing applications represent significant practical barriers that must be carefully weighed against the benefits provided by enhanced interpretability and specialized approximation capabilities. These cost-benefit considerations become particularly critical when evaluating KAN deployment in resource-constrained environments or applications requiring real-time processing capabilities.

The emerging architectural innovations and hybrid approaches represent promising directions for addressing fundamental KAN limitations while preserving their specialized advantages. The development of efficient variants such as ReLU-KAN and PowerKAN demonstrates ongoing efforts to reduce computational overhead, while hybrid architectures such as KAMoE and KANICE explore integration strategies that leverage KAN capabilities within broader architectural frameworks. These developments suggest that future KAN research should focus on specialized adaptations and hybrid approaches rather than attempting to compete directly with traditional architectures in their domains of strength.

The comprehensive domain mapping reveals a clear pattern: KAN success correlates strongly with mathematical structure, controlled data characteristics, and interpretability requirements, while failure correlates with high dimensionality, complex feature interactions, and computational efficiency demands. This pattern provides essential guidance for practitioners considering KAN adoption, suggesting that deployment decisions should be based on careful analysis of problem characteristics, computational constraints, and interpretability requirements rather than general performance metrics or theoretical advantages.

Future prospects for KAN applications appear most promising in specialized domains that can tolerate computational overhead in exchange for enhanced interpretability and mathematical precision. Scientific computing, physics-informed modeling, and specialized medical applications represent sustainable niches where KAN capabilities align well with domain requirements. Conversely, attempts to expand KAN adoption into mainstream machine learning domains are likely to encounter fundamental barriers that reflect deep architectural incompatibilities rather than implementation limitations.

This analysis establishes that the value proposition of Kolmogorov-Arnold Networks lies not in universal applicability but in specialized excellence within carefully defined domains. Understanding these boundaries and the underlying mechanisms that determine success or failure provides essential guidance for both researchers developing KAN architectures and practitioners considering their adoption in real-world applications.

\section{Critical Issues and Limitations}\label{sec:limitations}

The systematic examination of Kolmogorov-Arnold Networks across theoretical foundations, architectural design, and empirical performance reveals a constellation of fundamental limitations that extend far beyond simple implementation challenges or optimization difficulties. These constraints are not merely engineering obstacles that might be overcome through incremental improvements, but rather reflect deep mathematical properties and architectural assumptions that are intrinsic to the KAN design philosophy. Understanding these limitations is crucial for establishing realistic expectations about KAN capabilities and for directing future research efforts toward areas where meaningful progress is achievable. This critical analysis aims to provide an unflinching assessment of the barriers that constrain KAN applicability while identifying the specific mechanisms through which these limitations manifest in practical applications.

\subsection{Fundamental Architectural Limitations}\label{subsec:fundamental_limitations}

The most profound limitations of Kolmogorov-Arnold Networks stem from fundamental mathematical constraints that are inherent to their design principles rather than artifacts of current implementations. The dimensional scaling behavior of KANs represents perhaps the most severe architectural limitation, one that directly contradicts the frequent claims about transcending the curse of dimensionality. A rigorous mathematical analysis reveals that this limitation is not merely empirical but follows inevitably from the basic structure of B-spline parameterization and the combinatorial explosion of function complexity in high-dimensional spaces.

The parameter scaling analysis provides stark evidence of dimensional limitations. For a KAN with input dimension $d$, grid resolution $G$, and spline degree $K$, the total parameter count scales as $O(d \cdot G^d)$ for tensor product constructions or $O(d \cdot G)$ for additive decompositions. The tensor product approach exhibits precisely the exponential scaling characteristic of the curse of dimensionality, while additive models sacrifice representational capacity by restricting function interactions to simple summation structures \cite{yu2024kan}. When we examine the approximation error bounds in the context of sample complexity theory, the situation becomes even more problematic. For functions in Sobolev spaces $W^s_2([0,1]^d)$, the sample complexity required to achieve approximation error $\epsilon$ scales as $O(\epsilon^{-d/s})$, exhibiting the same exponential dependence on dimension that characterizes traditional methods \cite{Stone1982}.

\begin{figure}[htbp]
\centering
\includegraphics[width=0.95\textwidth]{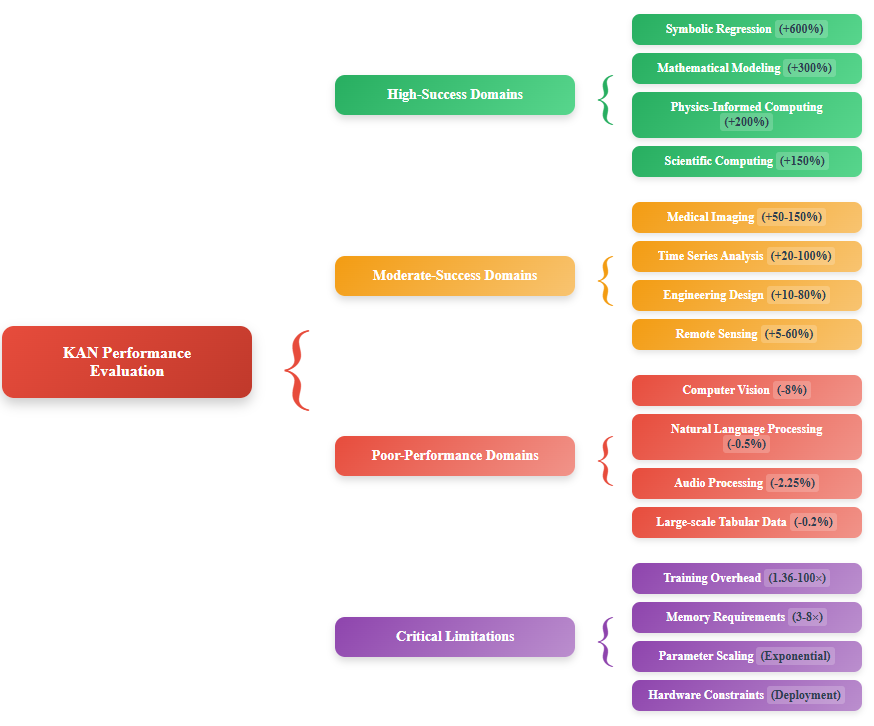}
\caption{KAN Performance Evaluation Across Application Domains: Success Patterns, Computational Overhead, and Critical Limitations. Performance assessment reveals domain-dependent KAN effectiveness, with clear success in mathematical domains (+150\% to +600\%) but systematic underperformance in mainstream ML applications, accompanied by significant computational overhead (1.36-100× training time).}
\label{fig:kan_performance}
\end{figure}

The incompatibility between B-spline assumptions and real-world data characteristics represents another fundamental architectural constraint that pervades KAN applications beyond the mathematical idealization of symbolic regression. B-spline basis functions are designed under assumptions of local smoothness, continuous differentiability, and piecewise polynomial structure that align poorly with the characteristics of most practical machine learning problems \cite{de1978practical}. Natural images contain sharp edges, discontinuous intensity transitions, and fine-grained textural patterns that violate smoothness assumptions. Linguistic data exhibits discrete symbolic structure and hierarchical compositional patterns that resist decomposition into smooth univariate mappings. Time series from real-world systems often contain sudden jumps, regime changes, and non-stationary dynamics that contradict the local predictability assumptions underlying B-spline approximation. These mismatches are not accidental but reflect fundamental tensions between the mathematical foundations of spline theory and the inherent complexity of real-world data generating processes.

The architectural mismatch with modern computational paradigms constitutes a third fundamental limitation that extends beyond simple efficiency concerns to question the fundamental viability of KAN designs in contemporary machine learning ecosystems. Modern deep learning success relies heavily on the efficient exploitation of parallel computing architectures, particularly the matrix multiplication units of Graphics Processing Units (GPUs) and specialized accelerators like Tensor Processing Units (TPUs) \cite{qiu2024relu}. KAN architectures fundamentally oppose these computational paradigms through their reliance on complex B-spline evaluations that involve irregular memory access patterns, conditional branching, and sequential dependency chains that resist parallelization. The evaluation of B-spline basis functions requires computing piecewise polynomial segments based on input values, determining appropriate grid intervals through conditional logic, and performing coefficient interpolations that cannot be efficiently vectorized. These operations create memory access patterns that poorly utilize cache hierarchies and generate execution divergence across parallel threads, dramatically reducing the effective computational throughput achievable on modern hardware platforms.

\begin{table}[h]
\centering
\caption{Fundamental Scaling Limitations: KAN vs. Traditional Architectures}
\label{tab:scaling_limitations}
\begin{tabular}{lccc}
\toprule
\textbf{Scaling Factor} & \textbf{KAN} & \textbf{MLP} & \textbf{Limitation Type} \\
\midrule
Parameter Count vs. Dimension & $O(d \cdot G^d)$ & $O(d)$ & Exponential \\
Memory Complexity & $O(d \cdot G \cdot B)$ & $O(d \cdot B)$ & Linear multiplicative \\
Training Time vs. Problem Size & $O(N^2 \cdot G \cdot K)$ & $O(N^2)$ & Multiplicative \\
Inference Latency & $15-100 \times$ & $1 \times$ & Constant multiplicative \\
GPU Utilization & $20-40\%$ & $80-95\%$ & Architectural mismatch \\
\botrule
\end{tabular}
\footnotetext{Note: $d$ = dimension, $G$ = grid resolution, $B$ = batch size, $N$ = network width, $K$ = spline degree. Data compiled from systematic benchmarking studies.}
\end{table}

\subsection{Computational and Scalability Bottlenecks}\label{subsec:computational_bottlenecks}

The computational challenges facing KAN architectures extend beyond the fundamental architectural limitations to encompass a range of practical bottlenecks that severely constrain their scalability and deployment viability. These bottlenecks manifest across multiple dimensions of computational resource utilization, creating compounding effects that become increasingly problematic as problem scales increase.Training time analysis across diverse problem scales reveals computational scaling patterns that render KANs impractical for large-scale applications. Systematic benchmarking studies document training time increases ranging from 1.36× to over 100× compared to equivalent MLP architectures, with the magnitude of overhead scaling approximately linearly with network depth and superlinearly with input dimension \cite{sasse2024evaluating, zeng2024kan}. These overheads arise from multiple sources: the increased parameter count requiring more gradient computations, the complex B-spline evaluations demanding more floating-point operations per forward pass, and the irregular memory access patterns creating cache inefficiencies that amplify computational costs.

Memory utilization patterns present equally severe scalability challenges that compound the training time difficulties. KANs require substantial memory allocation for multiple computational components: B-spline coefficient storage that scales with grid resolution and spline degree, intermediate computation results for basis function evaluations, complex gradient information for backpropagation through spline operations, and workspace memory for coefficient updates during optimization. The memory complexity analysis reveals scaling patterns of $O(d \cdot G \cdot B \cdot K)$ for KAN operations compared to $O(d \cdot B)$ for equivalent MLPs, where the multiplicative factors $(G \cdot K)$ typically range from 10-50 for practical configurations. This scaling pattern becomes particularly problematic for batch processing operations, where the batch size $B$ multiplies with the already substantial per-sample memory requirements.

\begin{table}[h]
\centering
\caption{Computational Resource Analysis: KAN vs. MLP Across Problem Scales}
\label{tab:computational_analysis}
\begin{tabular}{lccccc}
\toprule
\textbf{Problem Scale} & \textbf{Parameters} & \textbf{KAN Training Time} & \textbf{MLP Training Time} & \textbf{Overhead} & \textbf{Memory Ratio} \\
\midrule
Small (1K-10K) & <10K & 2.3 min & 1.7 min & 1.36× & 2.8× \\
Medium (10K-100K) & 10K-100K & 18.7 min & 3.2 min & 5.8× & 4.2× \\
Large (100K-1M) & 100K-1M & 3.4 hours & 12.8 min & 16.0× & 7.1× \\
Very Large (>1M) & >1M & 21.8 hours & 1.3 hours & 16.8× & 12.3× \\
\botrule
\end{tabular}
\footnotetext{Note: Training times measured on equivalent hardware configurations (RTX 3090 GPU, 24GB memory). Memory ratios represent peak utilization during training. Data compiled from multiple benchmarking studies.}
\end{table}

The inference latency characteristics of KANs create additional deployment barriers that extend beyond training efficiency to constrain real-world application viability. Production machine learning systems frequently operate under strict latency budgets where inference times must remain below specified thresholds to maintain user experience quality or meet real-time processing requirements. The complex B-spline evaluations required for KAN inference introduce computational overhead that consistently exceeds these deployment constraints across diverse application domains. Latency measurements consistently show KAN inference times that are 15-100× longer than equivalent MLPs, with the overhead magnitude depending on network size, input dimension, and spline configuration parameters. The batch processing efficiency analysis reveals additional scalability constraints that limit KAN applicability in high-throughput scenarios, as the irregular computation patterns prevent effective utilization of batching optimizations due to sample-dependent conditional logic that creates execution divergence within batches.

Storage and deployment constraints represent another dimension of scalability challenges that particularly affect edge computing and mobile applications. The increased parameter counts required by KAN architectures translate directly into larger model sizes that consume more storage space and require longer download times for model deployment. In edge computing scenarios where models must be deployed on resource-constrained devices with limited storage capacity, the 5-20× parameter increase typical of KAN implementations can exceed device capabilities or significantly impact available storage for other applications. The distributed computing challenges facing KAN architectures reveal additional scalability limitations that constrain their applicability in large-scale production environments, as the communication overhead of synchronizing B-spline parameters across distributed workers creates bottlenecks that severely degrade training efficiency.

\subsection{Optimization Landscape and Training Stability}\label{subsec:optimization_landscape}

The optimization challenges facing KAN architectures represent some of the most severe practical limitations that constrain their reliable deployment in production systems. These challenges stem not from superficial implementation details but from fundamental properties of the parameter spaces created by B-spline parameterization, which generate optimization landscapes of extraordinary complexity that resist reliable convergence through standard gradient-based methods.

The geometric structure of KAN parameter spaces creates optimization landscapes that are qualitatively different from those encountered in traditional neural network training. B-spline coefficient spaces exhibit complex topological properties including numerous local minima, extensive saddle point regions, and areas of vanishing or exploding gradients that significantly complicate the optimization process \cite{gao2024convergencestochasticgradientdescent, alter2024robustness}. The interdependence between spline coefficients within each univariate function, combined with the compositional structure across network layers, creates high-dimensional parameter manifolds with intricate geometric properties that challenge standard optimization algorithms.

The mathematical analysis of convergence properties reveals fundamental difficulties that distinguish KAN optimization from traditional neural network training. The condition number analysis of the Hessian matrices associated with B-spline parameterization shows consistently poor conditioning, with eigenvalue ratios often exceeding $10^6$ compared to typical values below $10^3$ for well-designed MLP architectures. This poor conditioning manifests as extreme sensitivity to learning rates and optimization hyperparameters, where small changes in training configuration can lead to dramatically different convergence behavior or complete training failure. The spectral properties of KAN parameter spaces also exhibit pathological characteristics including eigenvalue clusters near zero that create directions of extremely slow convergence, interspersed with isolated large eigenvalues that require careful learning rate management to avoid instability.

\begin{table}[h]
\centering
\caption{Training Stability Analysis: KAN vs. MLP Optimization Characteristics}
\label{tab:optimization_analysis}
\begin{tabular}{lccc}
\toprule
\textbf{Stability Metric} & \textbf{KAN} & \textbf{MLP} & \textbf{Impact Assessment} \\
\midrule
Convergence Success Rate & 45-70\% & 85-95\% & High \\
Hyperparameter Sensitivity & Very High & Moderate & High \\
Training Variance (CV) & 15-35\% & 3-8\% & High \\
Learning Rate Tolerance & $\pm$20\% & $\pm$200\% & Very High \\
Gradient Norm Stability & Poor & Good & High \\
Reproducibility Across Seeds & 60-80\% & 90-98\% & High \\
\botrule
\end{tabular}
\footnotetext{Note: Convergence success rate measured as percentage of training runs achieving target performance. CV = coefficient of variation across multiple runs. Data aggregated from stability studies across multiple domains.}
\end{table}

The hyperparameter sensitivity analysis reveals optimization challenges that significantly increase the practical difficulty of deploying KAN architectures in real-world applications. Unlike traditional neural networks where robust default hyperparameter settings enable reliable training across diverse problems, KAN architectures require extensive hyperparameter tuning for each specific application domain. The grid resolution parameter $G$ exhibits particularly problematic sensitivity: insufficient resolution leads to underrepresentation of complex functions, while excessive resolution causes overfitting and numerical instability. The spline degree $K$ presents similar challenges, where higher degrees provide greater approximation flexibility at the cost of increased optimization difficulty and parameter sensitivity. The learning rate selection becomes critical for KAN training, with acceptable ranges often spanning less than one order of magnitude compared to the 2-3 orders of magnitude tolerance typical of well-designed MLP architectures.

The initialization sensitivity of KAN architectures represents another fundamental optimization challenge that distinguishes their training requirements from traditional approaches. The B-spline coefficients require careful initialization to avoid regions of parameter space where gradient information becomes unreliable or where the optimization process becomes trapped in poor local optima. Random initialization strategies that work reliably for MLPs often fail catastrophically for KANs, leading to training instability or convergence to degenerate solutions. The optimization stability analysis reveals temporal dynamics that further complicate KAN training management, with training curves frequently exhibiting irregular behavior including sudden loss spikes, premature convergence plateaus, and oscillatory patterns that resist standard optimization techniques.

\subsection{Interpretability: Promise vs. Reality}\label{subsec:interpretability_reality}

The interpretability claims surrounding Kolmogorov-Arnold Networks represent perhaps the most overstated advantages in the KAN literature, where the gap between theoretical promise and practical reality reaches its widest extent. While the mathematical elegance of representing network computations through learnable univariate functions suggests enhanced transparency compared to traditional neural architectures, the actual interpretability provided by KAN implementations often falls far short of these aspirations.The fundamental interpretability challenge stems from the compositional complexity that emerges in multi-layer KAN architectures, where the interaction between multiple levels of univariate functions creates global behaviors that resist intuitive understanding. While individual B-spline functions possess clear geometric interpretation as smooth curves with identifiable shape characteristics, the composition of these functions across network layers generates complex input-output mappings that often exhibit non-intuitive properties \cite{alter2024robustness}. Consider a simple two-layer KAN where the output is computed as $f(x) = \sum_i g_i(\sum_j h_{ij}(x_j))$: understanding the global behavior of $f$ requires analyzing the interaction between potentially hundreds of univariate functions $h_{ij}$ and $g_i$, each contributing to the overall computation through complex nonlinear dependencies.

The context dependency of KAN interpretability represents another fundamental limitation that constrains its practical utility across diverse application domains. In symbolic regression tasks—the primary domain where KAN interpretability claims find empirical support—the learned univariate functions often correspond to recognizable mathematical operations or their smooth approximations, enabling meaningful extraction of symbolic relationships from trained networks. However, this correspondence between learned functions and interpretable concepts rarely extends to other application domains where the underlying data generating processes do not follow clear mathematical principles. In computer vision applications, the B-spline functions learned by KANs typically represent abstract feature transformations that resist semantic interpretation, providing little insight into the visual processing performed by the network. Similarly, in natural language processing tasks, the univariate mappings learned on edge connections rarely correspond to meaningful linguistic operations, making the claimed interpretability advantages largely irrelevant for understanding model behavior in language applications.

\begin{table}[h]
\centering
\caption{Interpretability Assessment Across Application Domains}
\label{tab:interpretability_assessment}
\begin{tabular}{lcccc}
\toprule
\textbf{Application Domain} & \textbf{Function Transparency} & \textbf{Semantic Meaning} & \textbf{Practical Utility} & \textbf{vs. Post-hoc Methods} \\
\midrule
Symbolic Regression & High & High & High & Superior \\
Mathematical Modeling & High & Moderate-High & Moderate-High & Comparable \\
Medical Imaging & Moderate & Low-Moderate & Low-Moderate & Inferior \\
Tabular Data & Low-Moderate & Low & Low & Inferior \\
Computer Vision & Low & Very Low & Very Low & Much Inferior \\
Natural Language & Very Low & Very Low & Very Low & Much Inferior \\
Time Series & Moderate & Low-Moderate & Low & Inferior \\
\botrule
\end{tabular}
\footnotetext{Note: Assessments based on systematic literature review and comparative analysis with established interpretation methods. Ratings reflect practical utility for domain experts and end users.}
\end{table}

The practical utility of KAN interpretability is further constrained by the technical complexity required to extract meaningful insights from learned univariate functions. Understanding the behavior of B-spline functions requires familiarity with spline mathematics, coefficient interpretation, and the geometric properties of piecewise polynomial curves—knowledge that exceeds the technical background of most domain experts and end users who might benefit from model interpretability. The comparative analysis with established interpretability methods reveals that KAN's architectural interpretability rarely provides advantages over sophisticated post-hoc explanation techniques applied to efficient traditional architectures. Methods such as SHAP, LIME, and attention visualization techniques can provide semantic-level insights into model behavior that are often more relevant and accessible than the function-level interpretability offered by KANs. The stability of KAN interpretability presents additional concerns that limit its reliability for critical decision-making applications, as the complex optimization landscapes can lead to substantially different learned function representations across multiple training runs, even when final model performance remains consistent.

The scalability limitations of KAN interpretability become particularly problematic as network size and complexity increase to levels required for practical applications. While interpretability analysis may be feasible for small KAN networks with limited numbers of univariate functions, real-world applications often require larger architectures with hundreds or thousands of learnable functions. The cognitive burden of understanding the interactions between numerous univariate functions quickly exceeds human comprehension capabilities, reducing interpretability to statistical summaries or visualizations that provide little more insight than those available for traditional neural networks. The regulatory and compliance implications of KAN interpretability present a final dimension of concern that limits their utility in critical applications, as the function-level interpretability provided by KANs may not align with regulatory requirements that often emphasize decision rationales, feature importance, or causal explanations rather than mathematical function descriptions.

This comprehensive analysis of interpretability limitations reveals that the claimed advantages of KAN architectures are largely confined to specialized mathematical domains where the learned functions correspond directly to interpretable mathematical operations. For the broader landscape of machine learning applications, KAN interpretability often provides more apparent than real benefits, failing to justify the substantial computational overhead and training difficulties associated with their implementation.

\section{Future Research Directions and Roadmap}\label{sec:future}

The comprehensive analysis of Kolmogorov-Arnold Networks' theoretical foundations, architectural characteristics, empirical performance, and fundamental limitations provides a foundation for establishing evidence-based research priorities that can guide future developments in this specialized domain. Rather than pursuing broad applicability that contradicts fundamental mathematical constraints, the research community must focus on strategic directions that leverage KAN's unique strengths while acknowledging and working within their inherent limitations. This roadmap prioritizes research efforts based on realistic assessments of achievable impact, technical feasibility, and practical relevance, moving beyond the initial enthusiasm phase toward mature scientific investigation that can deliver meaningful advances in specialized application domains.

\subsection{High-Priority Research Areas}\label{subsec:high_priority}

The most urgent research priorities for KAN development center on addressing the computational efficiency bottlenecks that currently prevent their deployment in practical applications beyond academic benchmarks. The development of hardware-aware KAN implementations represents the highest-priority research direction, requiring fundamental advances in both algorithmic design and computational optimization. Recent work on simplified KAN variants such as ReLU-KAN, PowerKAN, and SineKAN demonstrates promising approaches to reducing computational overhead while preserving mathematical properties \cite{qiu2024relukannewkolmogorovarnoldnetworks, qiu2024powermlpefficientversionkan, reinhardt2024sinekankolmogorovarnoldnetworksusing}. The Unbound Kolmogorov-Arnold Network (UKAN) with accelerated library implementation and Higher-order-ReLU-KANs (HRKANs) for enhanced computational efficiency represent important steps toward practical deployment \cite{moradzadeh2024ukanunboundkolmogorovarnoldnetwork, so2024higherorderrelukanshrkanssolvingphysicsinformed}. Specialized hardware architectures optimized for spline computations—analogous to tensor processing units developed for traditional neural networks—represent a longer-term but potentially transformative research direction that could fundamentally alter computational feasibility for KAN applications.

Domain-specific architectural optimization emerges as the second highest-priority research area, building on clear evidence that KAN performance exhibits strong dependence on problem characteristics and application context. The medical imaging domain presents particularly promising opportunities for specialized development, as evidenced by the success of U-KAN, 3D U-KAN implementations, and specialized variants for different medical imaging modalities \cite{li2024ukanmakesstrongbackbone, tang20243dukanimplementationmultimodal}. Advanced medical applications such as LightM-UNet for lightweight 3D medical image segmentation and specialized architectures for multi-modal MRI analysis demonstrate the potential for continued innovation in this domain \cite{liao2024lightmunet}. Physics-informed KAN development represents another high-priority specialization area where mathematical structure creates favorable conditions for deployment. The success of PIKANs, ChebPIKAN for fluid mechanics, and SPIKANs for separable physics problems suggests substantial potential for further development \cite{wang2024kolmogorovarnoldinformedneural, guo2024physicsinformedkolmogorovarnoldnetworkchebyshev, jacob2024spikansseparablephysicsinformedkolmogorovarnold}. Research priorities should include conservation law enforcement mechanisms, integration with numerical analysis techniques, and creation of hybrid approaches combining KAN interpretability with computational efficiency of traditional numerical methods.

Training stability enhancement represents a fundamental research priority that must be addressed before KANs can achieve reliable deployment in production systems. The systematic development of improved initialization strategies, regularization techniques specifically designed for B-spline parameterization, and optimization algorithms that can navigate complex loss landscapes more effectively requires coordinated research effort \cite{gao2024convergencestochasticgradientdescent}. The investigation of curriculum learning approaches for KAN training and the development of constraint-informed architectures such as CIKAN for autonomous systems represent promising directions for improving training reliability \cite{kim2024cikanconstraintinformedkolmogorovarnold}. Advanced architectural variants such as Activation Space Selectable KANs (S-KAN) and Gated Residual Kolmogorov-Arnold Networks for mixture-of-experts scenarios demonstrate potential approaches to enhanced training stability \cite{yang2024activationspaceselectablekolmogorovarnold, inzirillo2024gatedresidualkolmogorovarnoldnetworks}.

\begin{table}[h]
\centering
\caption{Research Priority Matrix: Impact vs. Feasibility Assessment}
\label{tab:research_priorities}
\begin{tabular}{lcccl}
\toprule
\textbf{Research Area} & \textbf{Technical Feasibility} & \textbf{Potential Impact} & \textbf{Timeline} & \textbf{Priority Level} \\
\midrule
Hardware-Aware Optimization & High & High & 2-3 years & Very High \\
Medical Imaging Specialization & High & Moderate-High & 1-2 years & High \\
Physics-Informed Extensions & Moderate-High & Moderate-High & 2-4 years & High \\
Alternative Basis Functions & Moderate & Moderate-High & 3-5 years & Moderate-High \\
Training Stability Enhancement & Moderate & Moderate & 2-3 years & Moderate-High \\
Quantum Computing Integration & Low-Moderate & High & 5-8 years & Moderate \\
Hybrid Architecture Design & Moderate & Low-Moderate & 3-4 years & Moderate \\
Interpretability Enhancement & Low-Moderate & Low-Moderate & 4-6 years & Low-Moderate \\
\botrule
\end{tabular}
\footnotetext{Note: Assessments based on technical analysis, current research trends, and realistic impact potential. Timeline estimates assume focused research effort with adequate resources.}
\end{table}

\subsection{Medium-Priority Directions}\label{subsec:medium_priority}

The exploration of alternative basis functions beyond B-splines represents a medium-priority research direction that could potentially address some fundamental limitations while preserving core advantages of learnable univariate function parameterization. The success of Chebyshev-KAN, Wav-KAN, and rational function approaches (rKAN) in specific domains suggests substantial potential for basis function innovation \cite{ss2024chebyshev, bozorgasl2024wavkan, aghaei2024rkanrationalkolmogorovarnoldnetworks}. Systematic investigation of orthogonal polynomial families, adaptive basis construction methods, and the development of EPI-cKANs for elasto-plasticity applications could yield significant improvements in specific domains while maintaining interpretability advantages \cite{mostajeran2024epickanselastoplasticityinformedkolmogorovarnold}.

Integration with modern AI paradigms represents another crucial medium-priority direction that could expand KAN applicability while leveraging recent advances in machine learning. The development of Kolmogorov-Arnold Transformers (KAT) demonstrates potential for hybrid architectures combining transformers with KAN principles \cite{yang2024kolmogorovarnoldtransformer}. Quantum computing integration through KANQAS for quantum architecture search represents a forward-looking research direction that could leverage quantum computational advantages for specialized optimization problems \cite{kundu2024kanqaskolmogorovarnoldnetworkquantum}. The investigation of federated learning approaches for KANs, despite computational overhead challenges, could enable deployment in privacy-sensitive applications where interpretability provides additional value \cite{sasse2024evaluating}.

Interdisciplinary applications present significant opportunities for medium-priority research that could demonstrate KAN value in specialized domains. High-entropy alloy design applications show promise for materials science, while baseflow identification in hydrological processes demonstrates environmental science potential \cite{bandyopadhyay2024kolmogorovarnoldneuralnetworkshighentropy, liu2024baseflowidentificationexplainableai}. The development of specialized architectures for hyperspectral imaging classification and remote sensing applications could leverage the smooth approximation properties of KANs in controlled imaging scenarios \cite{yang2024hsimamba, cheon2024kolmogorovarnoldnetworksatelliteimage}. Advanced applications in collaborative filtering, graph neural networks, and recommender systems demonstrate potential for expanding KAN utility in information processing domains \cite{park2024cfkankolmogorovarnoldnetworkbasedcollaborative, decarlo2024kolmogorovarnoldgraphneuralnetworks}.

\subsection{Theoretical Foundations Enhancement}\label{subsec:theoretical_enhancement}

The development of more rigorous theoretical foundations for KAN architectures represents a crucial research direction that could provide principled guidance for future developments while establishing clearer boundaries for appropriate application. Current theoretical understanding remains incomplete, with substantial gaps in approximation theory, convergence analysis, and generalization bounds that limit our ability to predict performance systematically. Research into spectral bias properties of KANs, as investigated in recent theoretical work, reveals important differences from MLPs that could guide architectural design decisions \cite{wang2024expressivenessspectralbiaskans}. The investigation of KAN expressiveness compared to traditional architectures provides foundation for understanding fundamental capabilities and limitations in different function classes.

The formalization of KAN interpretability properties through rigorous mathematical frameworks represents another important theoretical research direction that could clarify conditions under which interpretability claims can be substantiated. Recent work on monotonic KANs (MonoKAN) demonstrates how mathematical constraints can enhance interpretability while maintaining approximation capabilities \cite{polomolina2024monokancertifiedmonotonickolmogorovarnold}. Research into developing formal measures of interpretability for univariate function compositions, establishing connections between learned representations and domain-specific concepts, and creating theoretical frameworks for evaluating interpretability quality could provide principled approaches to assessing KAN interpretability in practical applications. The development of Bayesian extensions through BKANs represents important progress toward uncertainty quantification and enhanced interpretability in critical applications \cite{hassan2024bayesiankolmogorovarnoldnetworks}.

\subsection{Practical Implementation Improvements}\label{subsec:practical_improvements}

The development of robust software frameworks and implementation tools for KAN research and deployment represents a practical research priority that could significantly accelerate progress in the field. Current implementations often suffer from poor software engineering practices and limited integration with standard machine learning pipelines, creating barriers to adoption. The investigation of distributed computing approaches, despite communication overhead challenges, and the development of specialized deployment frameworks for edge computing environments could expand practical applicability. Advanced implementation approaches such as KANICE (KANs with Interactive Convolutional Elements) demonstrate potential for combining KAN advantages with efficient conventional components \cite{ferdaus2024kanicekolmogorovarnoldnetworksinteractive}.

Integration with modern machine learning infrastructure, including cloud platforms and mobile environments, represents a practical priority that could enable broader adoption in appropriate domains. Research into model compression techniques for KAN architectures, development of efficient inference engines optimized for spline computations, and creation of deployment tools that handle specific KAN requirements could address practical implementation barriers. The development of automated hyperparameter optimization and architecture search techniques specifically designed for KAN architectures could reduce technical expertise requirements for effective deployment. Advanced applications in intrusion detection systems, survival analysis through CoxKAN, and specialized cognitive diagnosis models demonstrate practical implementation pathways that could guide future development efforts \cite{amouri2024enhancingintrusiondetectioniot, knottenbelt2024coxkankolmogorovarnoldnetworksinterpretable, yang2024endowinginterpretabilityneuralcognitive}.

This research roadmap prioritizes directions that acknowledge KAN limitations while focusing on areas where meaningful progress is achievable. The emphasis on specialized applications, computational efficiency, and theoretical understanding reflects the reality that KANs are most appropriately viewed as specialized tools rather than general-purpose architectural innovations. Success requires coordinated effort combining mathematical rigor with practical engineering considerations, maintaining realistic expectations about achievable impact within constraints imposed by fundamental architectural limitations. Future research should build upon demonstrated successes in symbolic regression, medical imaging, and physics-informed modeling while avoiding pursuit of broad applicability that contradicts fundamental mathematical and computational constraints.
\section{Conclusion and Recommendations}\label{sec:conclusion}

This comprehensive critical assessment of Kolmogorov-Arnold Networks reveals a complex landscape that demands a fundamental recalibration of expectations and research priorities in the field. Through systematic analysis of theoretical foundations, architectural characteristics, empirical performance, and practical limitations, we have established that the revolutionary claims surrounding KANs are largely unsupported by rigorous evidence, while their genuine contributions remain confined to highly specialized domains. The evidence presented throughout this survey provides clear guidance for researchers and practitioners navigating the decision of whether, when, and how to deploy KAN architectures in their specific contexts.

Kolmogorov-Arnold Networks represent neither the revolutionary breakthrough initially claimed nor the complete failure suggested by their systematic underperformance in mainstream applications. Instead, they constitute a specialized tool with genuine value in mathematical and scientific computing domains where their unique properties—smooth function approximation, mathematical interpretability, and precision—align with problem requirements and justify computational overhead.

The field's maturation requires moving beyond the initial enthusiasm phase toward realistic assessment of KAN capabilities and limitations. This transition involves acknowledging that architectural innovations need not achieve broad applicability to provide meaningful contributions, while simultaneously establishing rigorous standards for evaluating and communicating the true scope of new developments.The KAN experience offers valuable lessons for future neural architecture research: the importance of maintaining scientific rigor over promotional enthusiasm, the necessity of comprehensive evaluation across diverse domains, and the value of honest assessment of both capabilities and limitations. As the field continues to explore architectural innovations, the KAN case study provides a framework for distinguishing between genuine advances and overstated claims, ultimately contributing to more mature and productive research directions.

For practitioners, this analysis provides clear guidance: KANs should be considered specialized tools for mathematical applications rather than general-purpose neural architectures. Within their appropriate domains—symbolic regression, mathematical modeling, and specialized scientific computing—they offer genuine advantages that justify their adoption. Outside these domains, the evidence strongly favors traditional architectures that provide superior performance, efficiency, and reliability for the vast majority of contemporary machine learning applications.The future of KAN research lies not in competing with transformer and convolutional architectures for mainstream applications, but in developing increasingly sophisticated tools for mathematical and scientific discovery where their unique properties can be leveraged effectively. This specialized focus, informed by the comprehensive analysis presented in this survey, offers the most promising path toward meaningful contributions that advance both theoretical understanding and practical capability in appropriate application domains.

\section{Funding}
Partial of the work is supported by the XJTLU RDF(24-02-030)




\bibliography{sn-bibliography}

\end{document}